\documentclass{article} 
\usepackage{iclr2025_conference,times}
\usepackage{wrapfig}
\usepackage{enumitem}
\usepackage{bm}
\usepackage{hyperref}      
\usepackage{url}            
\usepackage{graphicx,booktabs}  
\usepackage{amsmath, amssymb, amsfonts}  
\usepackage{nicefrac}       
\usepackage{microtype}      
\usepackage{subcaption}
\usepackage{natbib}
\usepackage{algorithmic,algorithm}
\usepackage{multicol,multirow,array}
\usepackage{enumerate}
\usepackage{tabularx}

\usepackage{amsmath}
\usepackage{amssymb}
\usepackage{mathtools}
\usepackage{amsthm}

\definecolor{light_red}{rgb}{1, 0.5, 0.5}
\definecolor{light_blue}{rgb}{0.5, 0.5, 1}
\definecolor{light_orange}{rgb}{1, 0.75, 0.5}
\definecolor{dark_green}{rgb}{0, 0.42, 0.3}
\definecolor{light_green}{rgb}{0.67, 0.88, 0.67}

\usepackage{amsmath,amsfonts,bm}

\def\eqref#1{equation~\ref{#1}}

\def\1{\bm{1}}

\DeclareMathAlphabet{\mathsfit}{\encodingdefault}{\sfdefault}{m}{sl}
\SetMathAlphabet{\mathsfit}{bold}{\encodingdefault}{\sfdefault}{bx}{n}

\usepackage{hyperref}
\usepackage{url}

\title{Refining Answer Distributions for Improved Large Language Model Reasoning}

\author{Soumyasundar Pal$^*$ \And  Didier Chételat$^*$ \And  Yingxue Zhang\thanks{Huawei Noah's Ark Lab, Canada, $^{\dagger}$McGill University, Mila, \& ILLS} \And Mark Coates$^{\dagger}$
}

\iclrfinalcopy 
\begin{document}

\maketitle
\vspace{-1.0em}
\begin{abstract}
Large Language Models (LLMs) have exhibited an impressive capability to perform reasoning tasks, especially if they are encouraged to generate a sequence of intermediate steps. 
Reasoning performance can be improved by suitably combining multiple LLM responses, generated either in parallel in a single query, or via sequential interactions with LLMs throughout the reasoning process.
Existing strategies for combination, such as self-consistency and progressive-hint-prompting, make inefficient usage of the LLM responses.
We present Refined Answer Distributions, a novel and principled algorithmic framework to enhance the reasoning capabilities of LLMs.
Our approach can be viewed as an iterative sampling strategy for forming a Monte Carlo approximation of an underlying distribution of answers, with the goal of identifying the mode -- the most likely answer. 
Empirical evaluation on several reasoning benchmarks demonstrates the superiority of the proposed approach.
\end{abstract}

\section{Introduction}\label{sec:introduction}
\vspace{-0.5em}
As Large Language Models (LLMs) have increased in size, they have demonstrated increasing reasoning abilities~\citep{brown2020language}, despite not being explicitly trained to reason~\citep{wei2022emergent}.
In particular, Chain-of-Thought (CoT) prompting has become standard for eliciting these abilities, either through few-shot examples~\citep{wei2022chain} or via a triggering sentence such as ``Let's think step by step."~\citep{kojima2022large}. Nevertheless, although LLMs often produce correct reasoning steps, they struggle with higher-level planning~\citep{saparov2022language}, motivating researchers to explore strategies to remedy this deficiency. An effective solution  is to sample several chains-of-thoughts and take the most common answer as the final vote, an approach called Self-Consistency (CoT+SC)~\citep{wang2023selfconsistency}.
However, despite its impressive empirical performance, the gains  quickly plateau on many benchmarks, often with no improvement after five samples~\citep{aggarwal2023adaptive}. Thus, a more complex reasoning strategy appears necessary.

One promising direction involves encouraging LLMs to iteratively refine their reasoning, like humans often do~\citep{shinn2024reflexion, li2023deliberate,gou2023critic,madaan2024self}.
However,~\citet{huang2023large} demonstrate that the capability and effectiveness of LLMs' self-correction is overstated in the existing literature due to the use of oracle labels for determining stopping criteria~\citep{shinn2024reflexion}, unfair experimental protocols~\citep{du2023debate}, and sub-optimal initial prompt design~\citep{madaan2024self}.
Moreover, the review/feedback prompts employed in these approaches are often long and complex, and include intricate, hand-crafted examples, tailored for specific domains or benchmarks.
In spite of such extensive prompt engineering,~\citet{huang2023large} observe that most of these approaches perform worse than self-consistency in a fair evaluation setting. 

In this paper, we propose a novel iterative strategy called {\em Refined Answer Distributions (RAD)} that offers a more principled and practical way of reasoning with refinement. We consider a setting where we can conduct LLM calls sequentially or in parallel. Our method does not make major changes to the initial prompt in later calls, and we do not need extensive prompting effort to invoke an LLM's review of the previous answers. The process starts by constructing an initial distribution of answers using CoT. In subsequent rounds, we incorporate the unique answers from the previous round in the prompt. This leads to a new collection of answers, which we use to refine the answer distribution via a marginalization process. Our approach is agnostic to the strategy for incorporating a previous answer in the prompt, provided that it satisfies a `probability flow' condition that we specify. In our numerical experiments, we show that an existing hint-based prompting strategy~\citep{zheng2023progressive} satisfies this condition for a broad spectrum of reasoning datasets.
By maintaining a distribution, we reduce sampling variance and make more efficient usage of the LLM calls.

We make the following contributions:
\begin{itemize}[leftmargin=*,itemsep=0em,topsep=0em]
\item We introduce a novel iterative refinement strategy for reasoning with LLMs, with the key differentiator that the method maintains and updates a {\em distribution} over answers.  Our work highlights that an LLM can indeed derive benefit from self-reflecting on distributions of its past answers when attempting reasoning tasks, without a need to resort to extensive prompt design or hand-crafted examples.
  
\item Via multiple experiments with GPT-3.5 Turbo~\citep{brown2020language}, GPT-4 Turbo~\citep{openai2024gpt4}, the cost efficient GPT-4o-mini, and Llama models~\citep{llama3_2024},  we show that our proposed approach leads to consistently improved reasoning performance compared to {\em state-of-the-art} baselines for the same number of LLM calls and comparable token cost. We conduct experiments carefully to ensure there is no evaluation bias in favour of methods that employ refinement. Notably, out of 36 experimental scenarios, we observe that the proposed RAD variants have the highest accuracy in 30.

\item We show in the experiments that
our approach is flexible in that it can be combined successfully with different strategies for obtaining an initial distribution of answers (e.g. Chain of Thoughts~\citet{wei2022chain}, Progressive Hint Prompting (PHP)~\citet{zheng2023progressive}).

\end{itemize}
\vspace{-0.5em}
\section{Problem Statement}
\vspace{-0.75em}
Let $x$ be a question or a task in natural language, described in one or more sentences. Its true answer is denoted $y$, which can take different forms depending on the context, such as a number, a True/False boolean variable, or an option (a)/(b)/(c) from a multiple-choice set.
Potentially, we also assume to have access to a (small) set of triplets $\mathcal{I}{=}\{(x_j, z_j, y_j)\}_{j=1}^K$ corresponding to semantically-similar questions $x_j$, answers $y_j$, and rationales $z_j$. Each rationale $z_j$ is a sequence of short sentences that describe the step-by-step reasoning process leading to the answer $y_j$.

We assume that we can query the LLM in series or in parallel. Our task is to design a strategy for prompting the LLM and combining the responses to provide an answer $\hat{y}$ for the question $x$.
Performance is measured in terms of the average accuracy of the response, i.e., $\mathbb{E}[\mathbf{1}(\hat{y} = y)]$ for the indicator function $\mathbf{1}$.
\vspace{-0.5em}
\section{Methodology}\label{sec:methodology}
\vspace{-0.75em}
When presented with the question, an LLM produces a random answer $\tilde{y}$, drawn from an internal distribution that is dependent on the prompt and the LLM's parameters.
To avoid notational clutter, we suppress these dependencies and denote this distribution by $p(\tilde{y}|x)$. 
This distribution is analytically intractable but one can sample from it directly by prompting the LLM and subsequently collecting its answer. 

The reasoning ability of the LLM, i.e., the probability of producing the correct answer, is improved by careful construction of the prompt. 
For example, an encouragement to produce an explanation/rationale in the form of a sequence of short sentences to describe the step-by-step reasoning process has been shown to ameliorate LLMs' performance significantly compared to direct prompting~\citep{wei2022chain}. We denote the provided rationale as $z$, so the response of the LLM is a pair $(z,\tilde{y})$. If rationale-annotated in-context examples are available, then reasoning can be improved by incorporating in the prompt a (small) set in the form of triplets $\mathcal{I}{=}\{(x_j, z_j, y_j)\}_{j=1}^K$. 

Viewing the LLM response as a sample from the distribution, we can hypothesize that, if the LLM is capable of effective reasoning for the presented question, the mode of the distribution is most likely to be the correct answer. We would therefore like to extract the mode. One approach is to sample, either in parallel or sequentially, multiple LLM responses (each containing a rationale and answer). We can then select the answer corresponding to the Monte Carlo estimate of the mode by taking a majority vote over the sampled responses~\citep{wang2023selfconsistency}.

It has been observed that LLM output can be improved via a refinement or self-reflection process~\citep{zheng2023progressive, wu2024get, li2023deliberate, madaan2024self, park2023generative}. 
In this process, the LLM is provided with its previous response and/or answer, and asked to take it into account, or criticize it, before producing a refined response. 

This observation is the cornerstone of our proposed methodology. Rather than seeking the mode of the original distribution $p(\tilde{y}|x)$, we construct a sequence of distributions $\{p_r(\tilde{y}|x)\}_{r>1}$, where each successive distribution in the sequence is constructed via a refinement process using samples from the previous distribution, as the number of interactions with the LLM, $r$, grows. 
This refinement process involves marginalization over the LLM's previous answers, which are refined in the current iteration. 
We initialize $p_{1}(\tilde{y}|x)=p(\tilde{y}|x)$, i.e., we start with the distribution of answers obtained from the first interaction with the LLM at $r=1$.
Our hypothesis is that the probability of the correct answer, $p_r(y|x)$, increases with $r$, so the mode of a distribution later in the sequence, i.e., $r>1$, is more likely to be correct than the mode of $p(\tilde{y}|x)$.

\begin{wrapfigure}{r}{0.625\textwidth}
\centering
\vspace{-2.5em}
\includegraphics[trim={20pt 25pt 10pt 18pt},clip, scale=0.655]
{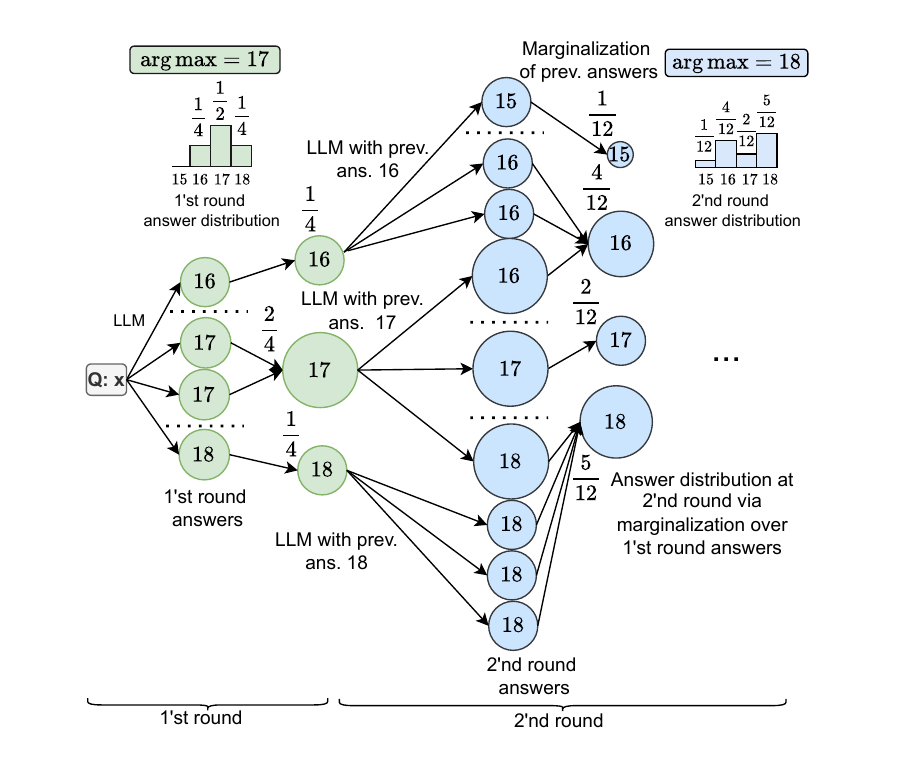}
\caption{Illustration of one iteration of our proposed method, Refined Answer Distribution (RAD). At its initialization, a distribution of answers is obtained from the LLM via multiple queries. In each subsequent iteration, new answers are sampled by refining each distinct old answer. The resulting samples are then accordingly weighted by the probability of the previous answers for marginalization.}
\label{fig:cot+hm}
\end{wrapfigure}
\vspace{-0.5em}
\subsection{Intuition}\label{subsec:intuit}
\vspace{-0.5em}
We now provide an example to illustrate why marginalizing over previous answers should make the mode of the inference distribution more likely to be the correct answer. Suppose that we are presented with a binary question $x$ with answer, say, $y=1$, and let us say that the probability of the correct answer is initially relatively low, $p(\tilde{y}{=}1|x)=0.4$. However, when we provide the correct answer on the prompt and ask the LLM to refine it, the LLM is much more likely to answer correctly, $p(\tilde{y}{=}1|x, \mathrm{Refine}(y{=}1))=0.8$. Refining the incorrect answer $0$ also strongly tilts the LLM towards that answer, but crucially, with slightly less probability, $p(\tilde{y}{=}0|x, \mathrm{Refine}(y{=}0))=0.6$. This is not unexpected or unusual, because it is often easier to see the truth of a statement in hindsight (or verify rather than solve unaided).  With our proposed marginalization procedure, the updated distribution of the answer would be:
\begin{align}
  p_{2}(\tilde{y}{=}1|x) &= p_{1}(\tilde{y}{=}1|x) p(\tilde{y}{=}1|x, \mathrm{Refine}(y{=}1)) + p_{1}(\tilde{y}{=}0|x) p(\tilde{y}{=}1|x, \mathrm{Refine}(y{=}0)) \,,\nonumber\\
&= 0.4\times 0.8 + (1-0.4)\times (1-0.6) = 0.56 > 0.4\,.\nonumber
\end{align}
Not only is the probability higher than before, but crucially, the mode of the distribution now aligns with the right answer ($y=1$).

 More generally, this augmentation will be observed \textbf{if and only if} the flow of probability mass into the correct answer exceeds the flow of probability mass out of the correct answer. The flow out is $p_{1}(\tilde{y}{=}y|x)\big(1-p(\tilde{y}{=}y|x, \mathrm{Refine}(y))\big)$, whereas the flow in is $\sum_{y' \neq y} p_{1}(\tilde{y}{=}y'|x) p(\tilde{y}{=}y|x, \mathrm{Refine}(y'))$. Since we expect $p_{1}(\tilde{y}{=}y|x, \mathrm{Refine}(y))$ to be close to 1, the flow out is likely to be small. By contrast, we might anticipate that when the LLM is presented with an incorrect answer, it can often ignore it to a large extent. Let us assume that $p(\tilde{y}{=}y|x, \mathrm{Refine}(y')) > c p_1(\tilde{y}{=}y|x)$ for all $y'$ for some positive constant $c<1$. Then the flow in exceeds $c p_1(\tilde{y}{=}y|x) (1-p_1(\tilde{y}{=}y|x))$. Thus, if $p(\tilde{y}{=}y|x, \mathrm{Refine}(y))  > 1-c(1- p_1(\tilde{y}{=}y|x))$, the mass assigned to the correct answer will increase. For example, consider $p_1(\tilde{y}{=}y|x)) = 0.4$ and $c=0.3$. Then we need $p(\tilde{y}{=}y|x, \mathrm{Refine}(y)) > 1{-}0.3{\times}(1{-}0.4) = 0.82$. 

We also note that repeated application of this procedure is further advantageous, which motivates the iterative version of our algorithm.
We formalize this intuition into a general procedure and provide an algorithm for approximating these distributions next.

\subsection{Refined Answer Distributions}
\vspace{-0.5em}
Our approach updates the distribution of answers by marginalizing over the answers obtained at the previous iteration. We denote the conditional probability of yielding $\tilde{y}$ as the answer for task $x$ with a previous answer $y'$ by $p(\tilde{y}|x, \mathrm{Refine}(y'))$.
We define a sequence of distributions $\{p_{r}(\tilde{y}|x)\}_{r>1}$, where two successive distributions are related as follows:
\begin{align}
p_{r{+}1}(\tilde{y}|x) = \int p(\tilde{y}|x, \mathrm{Refine}(y')) p_{r}(y'|x)\,dy'\,.\label{eq:hm-theory} 
\end{align}
The integral is replaced by a sum when $\tilde{y}$ is discrete, e.g., for multiple-choice questions.

\paragraph{Implementation:} We now outline the steps for performing one iteration of RAD. 
As a concrete example, Figure~\ref{fig:cot+hm} illustrates the procedure of approximating $p_{2}(\tilde{y}|x)$ from $p_{1}(\tilde{y}|x)$ in detail.   
Since neither $p(\tilde{y}|x, \mathrm{Refine}(y'))$ nor $p_{r}(\tilde{y}|x)$ can be computed analytically, we need to resort to a Monte Carlo approach for estimating $p_{r+1}
(\tilde{y}|x)$.

Suppose, at the end of the $r$-th iteration, $p_{r}(\tilde{y}|x)$ is approximated as follows:
\begin{align}
p_{r}(\tilde{y}|x) &\approx \sum_{m=1}^{M} \omega^{m} \delta(\tilde{y}-y^{m})\,,\label{eq:hm-mc-r}
\end{align}
where $\delta(\cdot)$ is the Kronecker delta function, $\{y^{m}\}_{m=1}^{M}$ is the set of distinct answers, and  $\omega^{m}$ is the estimated probability of obtaining the answer $y^{m}$ under the distribution $p_{r}(\cdot|x)$.
For example, in Figure~\ref{fig:cot+hm}, at $r=1$, we have $M=3$ distinct answers $y^1=16, y^2=17,$ and $y^3=18$ with estimated probabilities $\omega^{1} = \frac{1}{4}, \omega^{2} = \frac{1}{2}$, and $\omega^{3} = \frac{1}{4}$. 
If the correct answer $y=18$, then the LLM's current answer $\hat{y}=17$, based on the estimated mode of $p_{1}(\tilde{y}|x)$, is incorrect.

Assuming a sampling budget of $B_{r+1}$, which denotes the maximally allowed number of answers to be sampled at the $(r{+}1)$-th iteration, we modify, for each $m=1,\dots,M$, the prompt by appending $y^{m}$ as the previous answer, and sample $\lfloor \frac{B_{r{+}1}}{M} \rfloor$ answers subsequently.
This forms the following Monte Carlo approximation:
\begin{align}
p(\tilde{y}|x, \mathrm{Refine}(y=y^{m})) &\approx \sum_{\ell=1}^{L_{m}} \bar{\omega}^{\ell,m} \delta(\tilde{y}-y^{\ell,m})\,.\label{eq:php-mc}
\end{align}
Here, $\{y^{\ell,m}\}_{\ell=1}^{L_{m}}$ are the $L_{m}$ distinct answers extracted from the $\lfloor \frac{B_{r{+}1}}{M} \rfloor$ answers and  $\bar{\omega}^{\ell,m}$ is the estimated probability of having $y^{\ell,m}$ as the answer conditioned on the previous answer $y^{m}$.
In Figure~\ref{fig:cot+hm}, the total budget for $r=2$, i.e. $B_2 = 9$, the number of distinct answers for different previous answers are $L_1=2, L_2=3$, and $L_3=1$. For the previous answer $y^1=16$, the estimated conditional probabilities of the answers $y^{1,1}=15$ and $y^{2,1}=16$ are $\bar{\omega}^{1,1}=\frac{1}{3}$ and $\bar{\omega}^{2,1}=\frac{2}{3}$ respectively. 

Using equations~\ref{eq:hm-mc-r} and~\ref{eq:php-mc}, we can approximate~\eqref{eq:hm-theory} as follows:
\begin{align}
p_{r{+}1}(\tilde{y}|x) &\approx \sum_{m=1}^{M}\sum_{\ell=1}^{L_m}  \omega^{m} \bar{\omega}^{\ell,m} \delta(\tilde{y}-y^{\ell,m})=\sum_{n=1}^{N} \bar{\omega}^{n} \delta(\tilde{y}-\bar{y}^{n})\,.\label{eq:hm-mc-r-plus-one} 
\end{align}
Here, $N$ is the number of distinct answers among $\{y^{\ell,m}\}_{\ell=1, m=1}^{L_{m}, M}$. 
The probability of having $\bar{y}^{n}$ as the answer is estimated as:
\begin{align}
\bar{\omega}^{n} =  \sum_{m=1}^{M}\sum_{\ell=1}^{L_{m}}  \omega^{m} \bar{\omega}^{\ell,m} \mathbf{1}(y^{\ell,m}=\bar{y}^n)\,.\label{eq:hm-mc-one-answer} 
\end{align}
From Figure~\ref{fig:cot+hm}, we observe that at $r=2$, we have $N=4$, the distinct answers are $\bar{y}^1 =15, \bar{y}^2 =16, \bar{y}^3 =17,$ and $\bar{y}^4 =18$. 
As shown in eq.~\ref{eq:hm-mc-one-answer}, the probability of obtaining $\bar{y}^4 =18$ is $\bar{\omega}^4 = (\frac{2}{4}\times\frac{1}{3}) + (\frac{1}{4}\times 1) = \frac{5}{12}$.
We observe that the probability of obtaining the correct answer is increased in one round of RAD.

We can stop this procedure by applying a variety of stopping criteria.
For example, we can stop (i) after a fixed number of iterations (when $r{>}R$); or (ii) based on a predefined sampling budget $B_{max}$ (when $r{>}R$, for $R$ such that $\displaystyle{\sum}_{p=1}^{R} B_p {\leqslant} B_{max} {<} \displaystyle{\sum}_{p{=}1}^{R{+}1} B_p$); or (iii) when the estimate of the mode of $p_{r}(y|x)$ remains the same for two successive iterations. 
Algorithm~\ref{alg:hm} in Appendix~\ref{sec:alg} provides a pseudocode description.

\paragraph{Discussion:}
Intuitively, refining some previous answers is more likely to lead to the correct answer than others for a given reasoning task.
Our framework naturally defines the usefulness of an answer $y'$ at the end of the $r$-th iteration by the probability $p_{r}(y'|x)$, and subsequently weights the answers generated by refinement of $y'$ at the $(r{+}1)$-th iteration by this value, while updating the distribution of answers in eq.~\ref{eq:hm-theory}.

We also note that the RAD framework is agnostic to the choice of prompts and is generally applicable to any advanced prompting technique, as those methods combined with SC can be used to initialize $p_{1}(\tilde{y}|x)$ for subsequent RAD iterations.
Our contribution is thus orthogonal to prompt engineering approaches. 
In our implementation of RAD in Section~\ref{sec:experiments}, we adopt the hint-based prompting strategy of~\citep{zheng2023progressive} for the refinement of previous answers.
Beyond the measurement of reasoning accuracy, we conduct a detailed empirical analysis across multiple benchmarks and LLMs (Figures~\ref{fig:gpt4o-mini-stacked},~\ref{fig:gpt3.5-stacked}, and~\ref{fig:gpt4-stacked}, Tables~\ref{tab:p_values} and~\ref{tab:percent_of_inc_prob}), which provides statistically significant evidence that the use of this prompt satisfies the `probability flow' criterion specified in Section~\ref{subsec:intuit}.

\section{Experimental Results}\label{sec:experiments}
\vspace{-0.5em}
\paragraph{Benchmarks:} We evaluate the proposed RAD algorithm on six arithmetic benchmarks: AddSub~\citep{hosseini2014}, MultiArith~\citep{roy2015}, SingleEQ~\citep{koncel-kedziorski2015}, SVAMP~\citep{patel2021}, GSM8K~\citep{cobbe2021gsm8k}, and AQuA~\citep{ling2017}.
AddSub and SingleEq contain easier problems, whereas the tasks in MultiArith, SVAMP, GSM8K, and AQuA are more challenging.
In addition, we conduct experiments on the MATH~\citep{hendrycksmath2021} dataset, which consists of a large collection of significantly more difficult mathematical questions of seven subcategories.
In order to demonstrate the general applicability of RAD beyond mathematical reasoning, we also consider two BIG-Bench Hard~\cite{suzgun2023bbh} tasks, namely Date Understanding and Object Tracking. 
More details of the datasets are deferred to Appendix~\ref{sec:benchmarks}.

\paragraph{Models:}
We use five different language models: GPT-3.5 Turbo~\citep{brown2020language}, which was fine-tuned using RLHF from a previous version (GPT-3), its upgraded version GPT-4 Turbo~\citep{openai2024gpt4}, the more recent cost-efficient GPT-4o-mini, and two Llama-based models,  Llama-3-8b-instruct and Llama-3-70b-instruct ~\citep{llama3_2024}. All three GPT models are closed-source, but can be publicly accessed using the OpenAI API\footnote{\url{https://openai.com/api}}.
The Llama models are open-source, although in practice, we used a commercial API service\footnote{\url{https://replicate.com/meta/meta-llama-3-70b-instruct}}.

\paragraph{Baselines and Experimental Setting:} We compare our approach to few-shot CoT~\citep{wei2022chain}, its combination with SC~\citep{wang2023selfconsistency}, PHP~\citep{zheng2023progressive}, and PHP+SC.
We refer to the proposed algorithm as CoT+RAD, since the same few-shot prompt as CoT is employed to initialize our approach.
For relatively cheaper LLMs, GPT-3.5 Turbo and GPT-4o-mini, we also consider another variant of our method called PHP+RAD, where the initial answer distribution is obtained from several PHP provided answers (i.e., PHP+SC). We also include known results of alternative iterative refinement methods on the same models and datasets, namely Self-Refine~\citep{madaan2024self}, CRITIC~\citep{gou2023critic}, repeated introspection (Self-Convinced prompting~\citep{zhang2023self}), Multi-Agent Debate ~\citep{du2023debate}, multi-agent multi-model round table conference 
(ReConcile~\citep{chen2023reconcile}), and verification methods such as Self-Verification~\citep{weng2023self-verify} and Forward-Backward reasoning 
(FOBAR~\citep{jiang2024fobar}). 
Computational budget limitations prevented us from running every possible combination of model, benchmark and competitor, especially since none of these works include results on GPT-4-Turbo and GPT-4o-mini. However,~\citet{huang2023large} thoroughly investigated many of these methods and found these approaches systematically inferior to a simple Self-Consistency baseline (CoT+SC), which is corroborated by our experimental results. 
For the MATH dataset, we compare our approach with a recently proposed multi-agent prompting technique MACM~\citep{lei2024macm}, which progressively performs each intermediate computational step (akin to a thought in CoT), verifies its correctness using code, and determines whether it can help in reaching the final answer via several agent interactions.
In order to avoid prohibitive token cost, we only use GPT-4o-mini for the MATH dataset.
Additionally, we restrict the use of Llama models to the arithmetic benchmarks.
We conduct our experiments on an Intel(R) Xeon(R) Gold 6140 CPU @ 2.30GHz.

For a fair comparison with CoT+SC, which requires sampling of multiple CoTs, we ensure that the proposed RAD uses a comparable number of CoTs. We use a total budget of $B_{max}{=}40$ sampled CoTs in two iterations of CoT+RAD, with $B_1{=}5$, $B_2{=}15$, and $B_3{=}20$. 
We allocate more CoTs to the later iterations ($r>1$), since we need to estimate $p(\tilde{y}|x, \textrm{Refine}(y'))$ for multiple values of $y'$. As we initialize $p_{1}(\tilde{y}|x)$ with CoT+SC, increasing the number of CoTs does not contribute substantially to improved performance at $r=1$~\citep{aggarwal2023adaptive}.
For PHP+RAD, we perform one iteration of marginalization with $B_1{=}20$ and $B_2{=}20$. 

For the CoT+SC algorithm, we sample exactly 40 CoTs to report performance, as in~\citet{wang2023selfconsistency}.
For PHP, generating one answer requires at least 2 interactions, but the exact number of CoTs cannot be known beforehand. Therefore, in order to ensure a fair comparison, we collect PHP answers in the PHP+SC algorithm until the total number of LLM calls matches that of CoT+RAD, which ensures that PHP+SC has an inference time comparable to that of CoT+RAD.
Except for CoT and PHP, which use greedy decoding, a temperature of 0.7 is used for all sampling based approaches, following the experimental settings of~\citet{wang2023selfconsistency} and \citet{zheng2023progressive}. 
The answer extraction and cleansing is carried out by following the same steps laid out by~\citet{kojima2022large}.
Additionally, for all datasets except AQuA (where the answers are multiple choice between A-E), we use a 3'rd decimal rounding off of LLM answers and `ground truth' before comparing them.
This fixes some questions in most of those five arithmetic datasets and the MATH dataset for all competing algorithms, (e.g. the
`true’ answer is 0.066666, but the LLM’s answer is 0.067), where the LLM’s answer is essentially
correct, but is declared incorrect due to a rounding error. 
A symbolic evaluation using latex2sympy2\footnote{\url{https://pypi.org/project/latex2sympy2/}} is carried out to determine the correctness of the final answer for the MATH dataset (e.g. $2x{+}7$ is equivalent to $7{+}2x$). 
We measure the accuracy of the answer as the performance metric.
CoT employs the same 4-shot prompt for AQuA and the same 8-shot prompt for other four arithmetic datasets, as designed by~\citet{wei2022chain}.
For the MATH dataset and the BBH tasks, we use the same prompts as~\citet{zheng2023progressive} and~\citet{suzgun2023bbh} respectively. 
PHP and PHP+SC also use the same base prompts to obtain the initial answer(s). 
Example prompts for all algorithms can be found in Appendix~\ref{sec:full-prompts}.

\paragraph{Results on Arithmetic Benchmarks:} We summarize the experimental results using the GPT models in Table~\ref{tab:result}. Results using the weaker Llama models can be found as Table~\ref{tab:result_llama} in Appendix~\ref{sec:llama}.
For each dataset and LLM, we conduct a Wilcoxon signed rank test between the top two algorithms and declare their difference statistically significant at the 5\% level.
As we use more recent versions of the GPT models than in the original articles of CoT+SC~\citep{wang2023selfconsistency} and PHP~\citep{zheng2023progressive}, the results are not directly comparable, but are broadly in line with their reported numbers. 
We observe that for all LLMs, with or without SC, PHP achieves higher accuracy than CoT prompting in most cases, demonstrating the advantage of using the LLMs' answers as hints.
The superior accuracy of CoT+SC compared to the greedy decoding of CoT for the majority of datasets showcases the strong empirical performance of SC, arising due to the consideration of diverse reasoning paths.
PHP+SC emerges as a close competitor to CoT+SC in most cases, although the relative accuracy gain compared to PHP is much lower, since PHP in itself is a strong baseline. Since PHP+SC does not consistently outperform CoT+SC, we can conclude that the incorporation of hints alone is insufficient to achieve better reasoning accuracy.

\begin{table}[t]
\vspace{-1.em}
\caption{Mean and standard error of accuracy (in \%) of few-shot arithmetic reasoning. The \textbf{highest} accuracy among all competing algorithms using the same LLM is marked in \textbf{bold} and is shown in {\color{red}\textbf{red}}, {\color{blue}\textbf{blue}}, and {\color{orange}\textbf{orange}} for {\color{red}\textbf{GPT-3.5 Turbo}}, {\color{blue}\textbf{GPT-4 Turbo}}, and {\color{orange}\textbf{GPT-4o-mini}} respectively. 
The \underline{second-best} accuracy in those cases is marked with an \underline{underline} and is shown in \underline{{\color{light_red}light red}}, \underline{{\color{light_blue}light blue}}, and \underline{{\color{light_orange}light orange}} respectively. The \textbf{highest} accuracy is marked with an asterisk if the difference from the \underline{second-best} accuracy is statistically significant.
}
\vspace{-0.5em}
\label{tab:result}
\centering
\scriptsize
\begin{tabular}{cccccccc}
\toprule 
\textbf{LLM} &\textbf{Algorithm}   &\textbf{AddSub}        &\textbf{MultiArith}         &\textbf{SingleEQ} &\textbf{SVAMP}        &\textbf{GSM8K}         &\textbf{AQuA} \\ \midrule[0.25ex]
{\multirow{8}{*}{\rotatebox[origin=c]{90}{{\color{red}\textbf{GPT-3.5 Turbo}}}}} 

&\textbf{CoT}  &\underline{{\color{light_red}91.4$\pm$1.4}} &97.8$\pm$0.6 &97.0$\pm$0.7 &81.9$\pm$1.2 &78.2$\pm$1.1 &58.3$\pm$3.1 \\

&\textbf{PHP}   &{\color{red}\textbf{91.6$\pm$1.4}}$^*$ &99.2$\pm$0.4 &97.6$\pm$0.7 &83.4$\pm$1.2 &83.2$\pm$1.0 &59.1$\pm$3.1\\

&\textbf{CoT+SC}   &91.1$\pm$1.4 &99.0$\pm$0.4 &97.6$\pm$0.7 &85.1$\pm$1.1 &83.2$\pm$1.0 &69.3$\pm$2.9 \\ 

&\textbf{PHP+SC}     &90.6$\pm$1.5 &98.8$\pm$0.4 &97.4$\pm$0.7 &83.3$\pm$1.2 &85.2$\pm$1.0 &64.2$\pm$3.0 \\

&\textbf{Self-Refine} &-   &- &- &- &75.1 &- \\

&\textbf{CRITIC} &-  &- &- &83.3 &78.2 &- \\
&\textbf{Self-Convinced} &79.3  &- &- &84.9 &81.5 &62.0 \\
&\textbf{Multi-Agent (Debate)} &-  &- &- &- &85.0$\pm$3.5 &- \\
&\textbf{ReConcile} &-  &- &- &- &85.3$\pm$2.2 &\underline{{\color{light_red}66.0$\pm$0.8}} \\
&\textbf{Self-Verification} &90.4  &97.4 &92.9 &83.1 &74.9 &60.6 \\
&\textbf{FOBAR} &89.4  &\underline{{\color{light_red}99.3}} &94.5 &{\color{red}\textbf{88.9}} &85.1 &62.6 \\
\cmidrule[0.025ex]{2-8}
&\textbf{CoT+RAD}  &{\color{red}\textbf{91.6$\pm$1.4}}$^*$ &{\color{red}\textbf{99.7$\pm$0.2}}$^*$ &\underline{{\color{light_red}98.0$\pm$0.6}} &\underline{{\color{light_red}86.2$\pm$1.1}} &\underline{{\color{light_red}87.5$\pm$0.9}} &{\color{red}\textbf{70.5$\pm$2.9}}$^*$ \\
&\textbf{PHP+RAD}  &\underline{{\color{light_red}91.4$\pm$1.4}} &\underline{{\color{light_red}99.3$\pm$0.3}} &{\color{red}\textbf{98.4$\pm$0.5}}$^*$ &85.9$\pm$1.1 &{\color{red}\textbf{88.6$\pm$0.9}}$^*$ &{\color{red}\textbf{70.5$\pm$2.9}}$^*$ \\
\midrule[0.25ex] \midrule[0.25ex]

{\multirow{5}{*}{\rotatebox[origin=c]{90}{{\color{blue}\textbf{GPT-4 Turbo}}}}} 

&\textbf{CoT}  &{\color{blue}\textbf{96.5$\pm$0.9}}$^*$ &98.3$\pm$0.5 &96.5$\pm$0.8 &92.3$\pm$0.8 &86.4$\pm$0.9 &83.9$\pm$2.3 \\

&\textbf{PHP}   &{\color{blue}\textbf{96.5$\pm$0.9}}$^*$ &\underline{{\color{light_blue}98.5$\pm$0.5}} &\underline{{\color{light_blue}97.4$\pm$0.7}} &93.3$\pm$0.8 &\underline{{\color{light_blue}91.4$\pm$0.8}} &83.9$\pm$2.3 \\

&\textbf{CoT+SC}   &\underline{{\color{light_blue}96.2$\pm$1.0}} &{\color{blue}\textbf{98.8$\pm$0.4}}$^*$ &97.0$\pm$0.8 &93.4$\pm$0.8 &88.5$\pm$0.9 &{\color{blue}\textbf{85.8$\pm$2.2}}$^*$ \\ 

&\textbf{PHP+SC} 
&95.9$\pm$1.0 &{\color{blue}\textbf{98.8$\pm$0.4}}$^*$ &96.9$\pm$0.8 &\underline{{\color{light_blue}93.9$\pm$0.8}} &91.1$\pm$0.8 &82.7$\pm$2.3 \\
\cmidrule[0.025ex]{2-8}
&\textbf{CoT+RAD}  
&{\color{blue}\textbf{96.5$\pm$0.9}}$^*$ &{\color{blue}\textbf{98.8$\pm$0.4}}$^*$ &{\color{blue}\textbf{98.6$\pm$0.5}}$^*$ &{\color{blue}\textbf{94.6$\pm$0.7}}$^*$ &{\color{blue}\textbf{94.6$\pm$0.6}}$^*$ &\underline{{\color{light_blue}84.3$\pm$2.3}} \\ 
\midrule[0.25ex] 
\midrule[0.25ex]

{\multirow{6}{*}{\rotatebox[origin=c]{90}{{\color{orange}\textbf{GPT-4o-mini}}}}} 

&\textbf{CoT}            &92.9$\pm$1.3 &98.8$\pm$0.4 &94.5$\pm$1.0 &93.5$\pm$0.8 &91.5$\pm$0.8 &78.7$\pm$2.5   \\

&\textbf{PHP}   &93.9$\pm$1.2 &98.8$\pm$0.4 &95.3$\pm$0.9 &93.6$\pm$0.8 &93.2$\pm$0.7 &78.7$\pm$2.6   \\

&\textbf{CoT+SC}         &92.9$\pm$1.3 &98.8$\pm$0.4 &95.1$\pm$1.0 &94.0$\pm$0.8 &\underline{{\color{light_orange}93.6$\pm$0.7}} &82.7$\pm$2.4   \\

&\textbf{PHP+SC}        &92.9$\pm$1.3 &98.8$\pm$0.4 &95.1$\pm$1.0 &93.4$\pm$0.8 &93.4$\pm$0.7 &84.3$\pm$2.3   \\

\cmidrule[0.025ex]{2-8}
&\textbf{CoT+RAD}  &\underline{{\color{light_orange}94.4$\pm$1.2}} &98.8$\pm$0.4 &\underline{{\color{light_orange}95.7$\pm$0.9}} &\underline{{\color{light_orange}94.1$\pm$0.7}} &{\color{orange}\textbf{94.3$\pm$0.6}}$^*$ &\underline{{\color{light_orange}84.6$\pm$2.3}}   \\
&\textbf{PHP+RAD}  &{\color{orange}\textbf{96.5$\pm$0.9}}$^*$ &98.8$\pm$0.4 &{\color{orange}\textbf{98.4$\pm$0.6}}$^*$ &{\color{orange}\textbf{94.3$\pm$0.7}}$^*$ &{\color{orange}\textbf{94.3$\pm$0.6}}$^*$ &{\color{orange}\textbf{85.0$\pm$2.2}}$^*$ \\
\bottomrule
\end{tabular}
\vspace{-1.5em}
\end{table}

Our approach, CoT+RAD, considerably outperforms CoT+SC in most cases.
The  PHP+RAD variant performs comparably to CoT+RAD on GPT-3.5 Turbo but shows improved performance on GPT-4o-mini. This shows that our RAD approach is generally applicable, as it can be combined with different prompting methods for initialization, and it is not overly sensitive to the choice of hyperparameters.

\begin{figure}[t]
\centering
\includegraphics[width=0.75\textwidth]{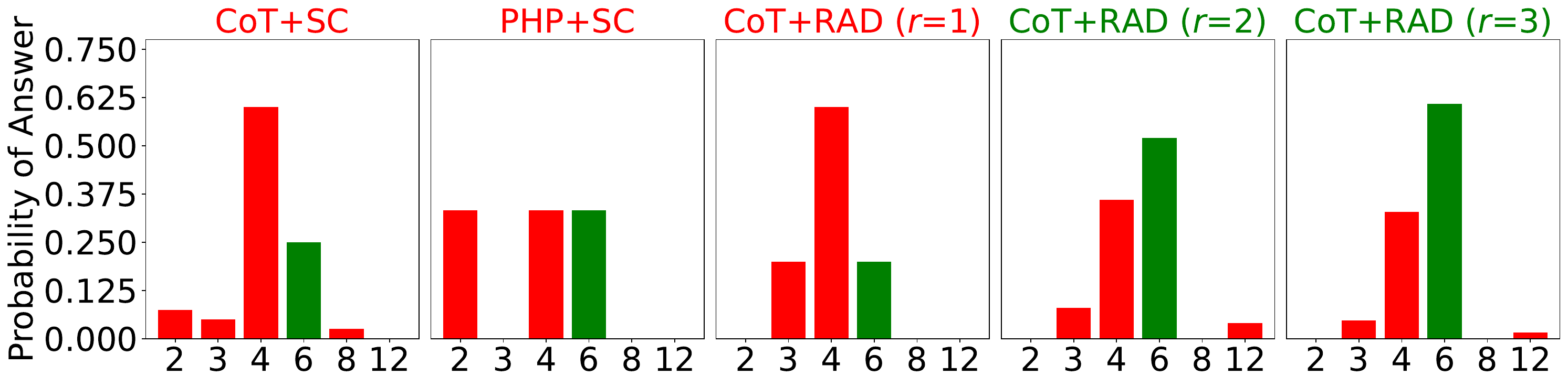}
\vspace{-0.5em}
\caption{The estimated probabilities of different answers from CoT+SC, PHP+SC, and CoT+RAD (using GPT-3.5 Turbo) for an example from GSM8K dataset.\\\textbf{Question:} The ice cream parlor was offering a deal, buy 2 scoops of ice cream, get 1 scoop free.  Each scoop cost \$1.50.  If Erin had  \$6.00, how many scoops of ice cream should she buy? \textbf{Answer:} {\color{dark_green}\textbf{6}}.}
\label{fig:gsm-example}
\vspace{-1.75em}
\end{figure}

\begin{figure}[t]
\centering
\vspace{-0.5em}
\includegraphics[trim={0em 33em 0em 4em }, clip, width=1\textwidth]{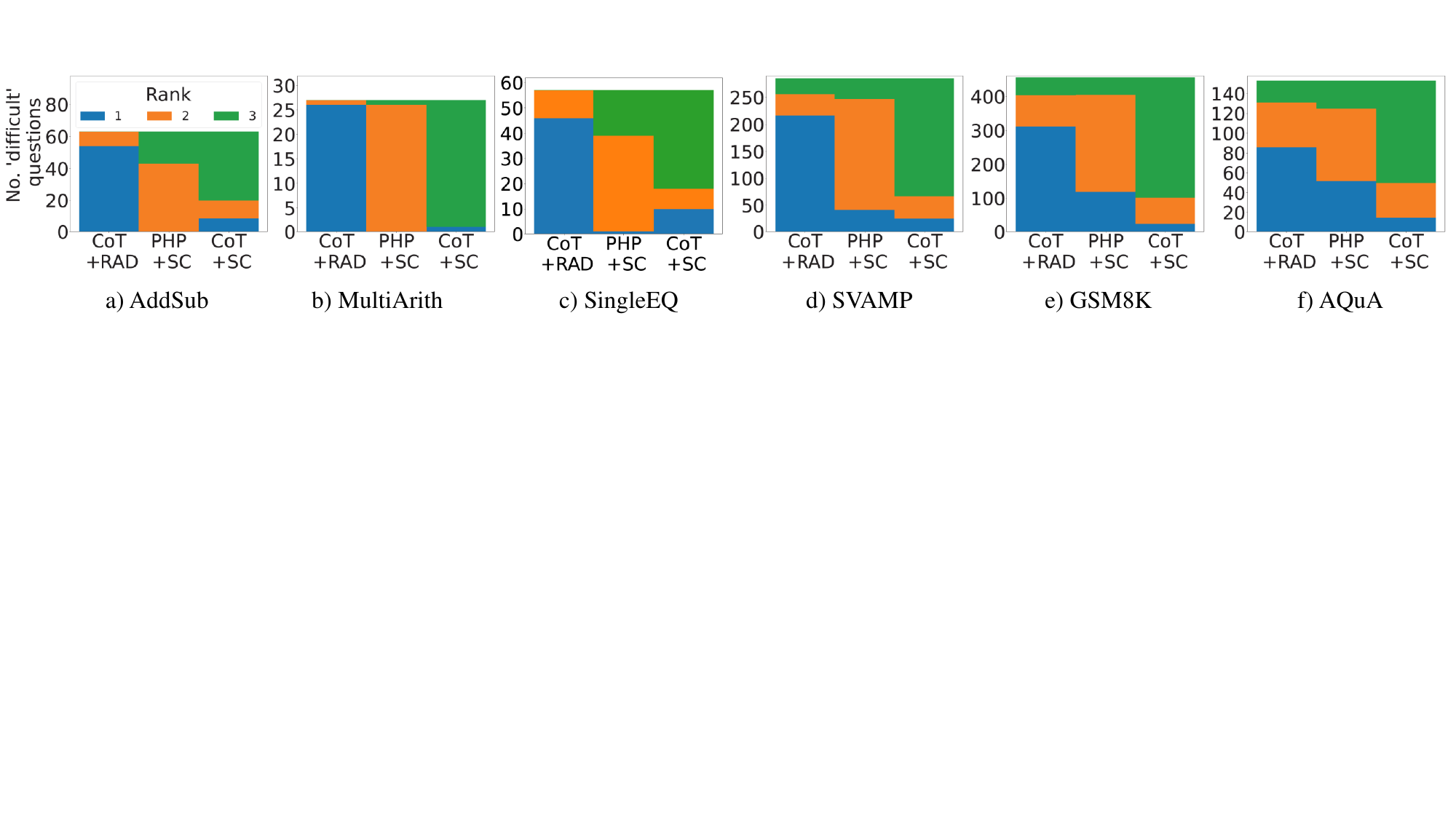}
\vspace{-1.75em}
\caption{Histogram of \textbf{ranks} of the algorithms (the \textbf{highest probability of the correct answer} results in the \textbf{lowest rank}) for the \textbf{`difficult'} questions from all six arithmetic datasets using \textbf{GPT-4o-mini}.}
\label{fig:gpt4o-mini-stacked}
\vspace{-1.5em}
\end{figure}

One benchmark that deviates from this pattern is AQuA using GPT-4 Turbo, where the best performing procedure is CoT+SC. 
This might be due to the fact that AQuA is the only multiple-choice question-answering benchmark among the six, and the employed hinting prompt ``The answer is close to A)" makes less sense for these types of questions. 
Further research on how to better extend PHP's hinting prompt to these types of problems might be valuable. 
In addition, all methods perform only as well as (or even worse than) a vanilla few-shot CoT and PHP on AddSub for both GPT-3.5 Turbo and GPT-4 Turbo models, possibly indicating the fact that the gains to be had using advanced methods on a dataset containing relatively simple questions are rather limited.

Figure~\ref{fig:gsm-example} shows the estimated probabilities of different answers of an example question from GSM8K for all sampling based algorithms using GPT-3.5 Turbo.
We observe that, while both CoT+SC and PHP+SC fail to reason correctly, the proposed CoT+RAD outputs the correct answer at both $r{=}2$ and $3$, although its initial distribution (computed using CoT+SC with $B_1{=}5$ samples) does not have a mode at the correct answer. 
More interestingly, CoT+SC cannot fix the error even if the budget increases to 40 from 5.
On the contrary, the proposed CoT+RAD utilizes the additional inference cost effectively to increase the probability of the correct answer at each iteration, demonstrating the usefulness of performing RAD in multiple iterations.

While Figure~\ref{fig:gsm-example} shows that CoT+RAD has a higher probability of the correct answer for a specific example question, a dataset-level investigation is necessary to determine whether this phenomenon is general. 
To that end, we restrict
ourselves to only the `difficult’ questions in these benchmarks.
If a question is correctly solved by all
algorithms in Table~\ref{tab:result}, we categorize it as `easy’. 
A question that is not `easy’ is termed `difficult’. 
All easy questions are
subsequently removed from the datasets\footnote{Since in each of these datasets, the majority of the questions are `easy', all of CoT+SC, PHP+SC, and CoT+RAD methods assign a very high probability on the correct answers for them.
In order to bring out the differences among these algorithms, we only focus on the `difficult' questions.}.
For all `difficult’ questions, we rank CoT+SC, PHP+SC, and CoT+RAD in terms of the probability they assign to the correct answer. 
The stacked-histograms of these ranks for all six datasets using GPT-4o-mini are shown in Figure~\ref{fig:gpt4o-mini-stacked}.
We observe that the proposed CoT+RAD achieves the lowest rank based on the probability of correct answer across all `difficult' questions for all datasets more often, outperforming both CoT+SC and PHP+SC.
This demonstrates that CoT+RAD has higher probability of the correct answer compared to its competitors for most of these `difficult' questions, which supports our intuition, presented in Section~\ref{subsec:intuit}.
Similar results are obtained for the other two LLMs (see Appendix~\ref{sec:prob_other_llm}).

\begin{table}[htbp]
\vspace{-0.5em}
\caption{Mean and standard error of accuracy (in \%) of reasoning on the MATH dataset using GPT-4o-mini. The \textbf{highest} accuracy among all competing algorithms is marked in \textbf{bold} and the \underline{second-best} accuracy in those cases is marked with an \underline{underline}. The \textbf{highest} accuracy is marked with an asterisk if the difference from the \underline{second-best} accuracy is statistically significant.}
\vspace{-0.75em}
\label{tab:result_math}
\centering
\scriptsize
\begin{tabular}{cccccccc}
\toprule
\textbf{Algorithm}    & \textbf{Algebra} & \begin{tabular}[c]{@{}c@{}}\textbf{Counting and}\\ \textbf{Probability}\end{tabular} & \textbf{Geometry} & \begin{tabular}[c]{@{}c@{}}\textbf{Intermediate}\\ \textbf{Algebra}\end{tabular} & \begin{tabular}[c]{@{}c@{}}\textbf{Number}\\ \textbf{Theory}\end{tabular} & \textbf{Prealgebra} & \textbf{Precalculus} \\ 
    \midrule
\textbf{CoT} &88.5$\pm$0.9      &73.4$\pm$2.0                                                                    &55.1$\pm$2.3          &51.5$\pm$1.6                                                                &76.3$\pm$1.8                                                        &86.9$\pm$1.1            &49.1$\pm$2.1            \\
\textbf{PHP} &90.2$\pm$0.9         &75.3$\pm$2.0                                                                    &55.9$\pm$2.3          &52.3$\pm$1.7                                                                &78.1$\pm$1.8                                                         &87.6$\pm$1.1           &51.1$\pm$2.1  

\\
\textbf{MACM} &90.8$\pm$0.9         &76.4$\pm$2.0                                                                   &57.4$\pm$2.3          &55.5$\pm$1.7                                                                &81.9$\pm$1.7                                                         &87.8$\pm$1.0           &\underline{51.3$\pm$2.1} 

\\
\textbf{CoT+SC}  &\underline{93.9$\pm$0.7}         &\textbf{82.9$\pm$1.7}$^*$                                                                    &\underline{64.7$\pm$2.2}          &\underline{58.1$\pm$1.7}                                                                &\underline{83.5$\pm$1.6}                                                          &\textbf{91.2$\pm$1.0}$^*$            &\underline{51.3$\pm$2.1}      \\
\midrule[0.25ex]
\textbf{PHP+RAD}  &\textbf{94.8$\pm$0.6}$^*$ &\underline{80.6$\pm$1.8} &\textbf{65.3$\pm$2.2}$^*$  &\textbf{58.9$\pm$1.6}$^*$ &\textbf{85.4$\pm$1.5}$^*$ &\underline{90.7$\pm$1.0} &\textbf{52.0$\pm$2.1}$^*$    \\
\bottomrule        
\end{tabular}
\vspace{-2em}
\end{table} 

\paragraph{Results on the MATH Dataset:} Table~\ref{tab:result_math} summarizes the experimental results for the MATH dataset, which is a large collection of significantly challenging mathematical reasoning problems.
For several sub-disciplines (Geometry, Intermediate Algebra, Precalculus), the state-of-the-art performance (without using extreme computation and a very long inference time) is in the range of 50-65 percent, which suggests that LLMs still find these problems very difficult to solve.
Since PHP outperforms CoT for all subcategories, we only evaluate PHP+RAD on these datasets to reduce the token cost, anticipating that PHP+SC would provide better initialization for RAD compared to CoT+SC.
Using GPT-4o-mini, the API cost of proposed PHP+RAD is approximately 2.9 cents on average, which is a modest increase from 2.5 cents of CoT+SC.
On the contrary, MACM incurs a significantly increased cost of approximately 6.4 cents, due to repeated LLM calls to perform and verify each step and utilization of code-interpreter.

We observe that despite performing an extensive segmentation of the reasoning task and code-based verification of each step, MACM has significantly lower accuracy compared to CoT+SC, which demonstrates that sophisticated prompting approaches often fail to outperform much simpler techniques in a fair experimental setting. 
The proposed PHP+RAD algorithm leads to a performance improvement in 5 out of 7 settings.

\paragraph{Big-Bench Hard Tasks:}
\begin{wraptable}{r}{0.5\textwidth}
\vspace{-3.0em}
\caption{Mean and standard error of accuracy (in \%) of reasoning for Date Understanding and Object Tracking tasks using GPT-4o-mini. The \textbf{highest} accuracy among all competing algorithms is marked in \textbf{bold} and the \underline{second-best} accuracy in those cases is marked with an \underline{underline}.}
\vspace{-0.75em}
\label{tab:result_other}
\centering
\scriptsize
\begin{tabular}{cccc}
\toprule
\textbf{Algorithm}
    & \begin{tabular}[c]{@{}c@{}}\textbf{Date}\\  \textbf{Understanding}\end{tabular} & \begin{tabular}[c]{@{}c@{}}\textbf{Object} \\ \textbf{Tracking}\end{tabular}   \\
\midrule    
\textbf{CoT} &91.9$\pm$1.4                                                               &96.4$\pm$0.7                                                                 \\
\textbf{PHP} &93.5$\pm$1.3                                                               &\underline{97.7$\pm$0.5}                                                            &   \\
\textbf{CoT+SC} &\underline{93.8$\pm$1.3}                                                               &96.7$\pm$0.7                           \\
\midrule[0.25ex]
\textbf{CoT+RAD} &\textbf{94.6$\pm$1.2}$^*$                                                               &\textbf{98.0$\pm$0.5}                                                           \\

\bottomrule    \\     
\end{tabular}
\vspace{-6em}
\end{wraptable}

In Table~\ref{tab:result_other}, we provide results for Date Understanding and Object Tracking, which are problems sets involving quantitative (but not strictly mathematical or arithmetic) reasoning.
We observe that PHP still outperforms CoT, demonstrating the utility of refinement via hinting beyond arithmetic tasks.
The proposed CoT+RAD offers an improvement in accuracy over the baselines for both of these datasets.

\vspace{-0.75em}
\section{Related Work}\label{sec:literature}
\vspace{-0.75em}

Our proposed method can be situated within a larger literature that aims to improve LLMs' reasoning ability through iterative refinement of chains-of-thought. These works primarily differ in the strategy used to refine the reasoning. 

One strand of work involves attempting to iteratively improve single answers, rather than whole distributions of answers like in our work. Progressive Hint Prompting (PHP)~\citep{zheng2023progressive} proposes to repeatedly generate chains-of-thought, each time encouraging new answers to look like previous answers by providing them as `hints' to the LLM.
Similar works use the same process but push answers away, rather than closer, to the previous answer. Progressive Rectification Prompting \citep{wu2024get} uses a prompt of the form `The answer is likely not $<$hint$>$', whereas Deliberate-then-Generate \citep{li2023deliberate} assumes an error was committed and asks the LLM to identify and correct the mistake. 
Hint-before-Solving Prompting \citep{fu2024hintbeforesolving} also utilizes hints, but in the form of key ideas like a mathematical formula, rather than an answer value.

Instead of trying to improve answers through hints, several works have instead tried to do the same using verbal criticism, at the cost of increased complexity.
Self-Refine~\citep{madaan2024self} incorporates a prompt where the LLM self-criticizes its answer, before being queried again with this reflection. 
Generative Agents~\citep{park2023generative} use a similar procedure, albeit in the context of an agent interacting with an environment. 
CRITIC~\citep{gou2023critic} is a more general framework, where the criticism prompt can make use of external tools like a web search engine to offer grounded corrections. Self-Convinced Prompting~\citep{zhang2023self} and Reflexion \citep{shinn2024reflexion} expand on Self-Refine by adding extra modules such as a separate answer encoder, or separating the evaluation and self-reflection dimensions of criticism into separate modules.
Finally, other related approaches include multi-round debate~\citep{du2023debate} and consensus via weighted voting mechanism~\citep{chen2023reconcile}.

Recent studies have, however, cast doubt on the ability of LLMs to self-criticize effectively~\citep{huang2023large, tyen2023llms}, leading researchers to consider using a separately trained LLM as the critic. In general, these methods generate a sequence of chains-of-thought, whereas we propose to refine the {\em distribution} of answers. REFINER \citep{paul2023refiner} fine-tunes a separate critic by supervised learning on examples perturbed by hand-designed rules and GPT-3.5 Turbo.
Retroformer \citep{yao2023retroformer} and RL4F \citep{akyurek2023rl4f} consider fine-tuning of the critic using reinforcement learning instead, which allows for a more precise alignment with the task of improving answers.


Finally, our work can be seen within the greater context of trying to improve chain-of-thought reasoning within large language models.
In existing work, several directions for improving CoTs are considered, including construction of better prompts to aid the LLM in reasoning~\citep{fu2023, zhang2023automatic}, fine-tuning with CoTs~\citep{zelikman2022} so that the LLMs learn to reason, and effective exploration strategies for multi-hop reasoning~\citep{besta2023, yao2023}. A recent survey by~\citet{chu2023survey} provides a comprehensive overview of these techniques.
Our contribution is orthogonal to these prompting techniques since we consider improving the {\em distribution} of answers iteratively rather than focusing on individual CoTs. 
Novel variants of RAD can be constructed by using these methods for initialization.


\vspace{-0.75em}
\section{Conclusion}\label{sec:conclusion}
\vspace{-0.5em}
This work presents a novel algorithmic approach, Refined Answer Distributions, to enable an LLM to solve a reasoning task by iteratively refining its inference distribution. The proposed algorithm addresses the issue of the diminishing marginal utility of extra LLM calls for Self-Consistency. RAD focuses on the distribution over the answers at each stage and assigns weights to the previous answers accordingly, concentrating on promising candidates. The marginalization procedure improves sample efficiency. The experimental results, over a range of quantitative reasoning benchmarks and several LLM variants, provide strong evidence that the approach leads to improved reasoning for the same budget of LLM calls, compared to Self-Consistency and other state-of-the-art refinement approaches. 

The work could be extended in several directions. Our experiments focus on quantitative reasoning tasks, but the method applies to other types of tasks as long as an appropriate answer refinement strategy would be chosen. For example, in tasks that require a verbal response, the prompt could incorporate `verbal criticism', based on one of the approaches detailed in Section~\ref{sec:literature}. In addition, in the current version of the procedure, we assign the same number of LLM calls to each unique answer from the previous round. Investigating more efficient strategies to allocate LLM calls non-uniformly to different answers could be another worthwhile direction.

\bibliography{reference}

\begin{thebibliography}{43}
\providecommand{\natexlab}[1]{#1}
\providecommand{\url}[1]{\texttt{#1}}
\expandafter\ifx\csname urlstyle\endcsname\relax
  \providecommand{\doi}[1]{doi: #1}\else
  \providecommand{\doi}{doi: \begingroup \urlstyle{rm}\Url}\fi

\bibitem[Aggarwal et~al.(2023)Aggarwal, Madaan, Yang, and {Mausam}]{aggarwal2023adaptive}
P.~Aggarwal, A.~Madaan, Y.~Yang, and {Mausam}.
\newblock Let{'}s sample step by step: {A}daptive-consistency for efficient reasoning and coding with {LLM}s.
\newblock In \emph{Proc. Conf. Empirical Methods in Natural Language Process.}, 2023.

\bibitem[Aky{\"u}rek et~al.(2023)Aky{\"u}rek, Aky{\"u}rek, Madaan, Kalyan, Clark, Wijaya, and Tandon]{akyurek2023rl4f}
A.~F. Aky{\"u}rek, E.~Aky{\"u}rek, A.~Madaan, A.~Kalyan, P.~Clark, D.~Wijaya, and N.~Tandon.
\newblock {RL4F}: {G}enerating natural language feedback with reinforcement learning for repairing model outputs.
\newblock \emph{arXiv preprint arXiv:2305.08844}, 2023.

\bibitem[{Besta} et~al.(2023){Besta}, {Blach}, {Kubicek}, {Gerstenberger}, {Gianinazzi}, {Gajda}, {Lehmann}, {Podstawski}, {Niewiadomski}, {Nyczyk}, and {Hoefler}]{besta2023}
M.~{Besta}, N.~{Blach}, A.~{Kubicek}, R.~{Gerstenberger}, L.~{Gianinazzi}, J.~{Gajda}, T.~{Lehmann}, M.~{Podstawski}, H.~{Niewiadomski}, P.~{Nyczyk}, and T.~{Hoefler}.
\newblock Graph of thoughts: {S}olving elaborate problems with large language models.
\newblock \emph{arXiv e-prints arXiv:2308.09687}, 2023.

\bibitem[Brown et~al.(2020)Brown, Mann, Ryder, Subbiah, Kaplan, Dhariwal, Neelakantan, Shyam, Sastry, Askell, Agarwal, Herbert-Voss, Krueger, Henighan, Child, Ramesh, Ziegler, Wu, Winter, Hesse, Chen, Sigler, Litwin, Gray, Chess, Clark, Berner, McCandlish, Radford, Sutskever, and Amodei]{brown2020language}
T.~B. Brown, B.~Mann, N.~Ryder, M.~Subbiah, J.~Kaplan, P.~Dhariwal, A.~Neelakantan, P.~Shyam, G.~Sastry, A.~Askell, S.~Agarwal, A.~Herbert-Voss, G.~Krueger, T.~Henighan, R.~Child, A.~Ramesh, D.~M. Ziegler, J.~Wu, C.~Winter, C.~Hesse, M.~Chen, E.~Sigler, M.~Litwin, S.~Gray, B.~Chess, J.~Clark, C.~Berner, S.~McCandlish, A.~Radford, I.~Sutskever, and D.~Amodei.
\newblock Language models are few-shot learners.
\newblock In \emph{Proc. Adv. Neural Inf. Process. Syst.}, 2020.

\bibitem[{Chih-Yao Chen} et~al.(2023){Chih-Yao Chen}, {Saha}, and {Bansal}]{chen2023reconcile}
J.~{Chih-Yao Chen}, S.~{Saha}, and M.~{Bansal}.
\newblock Re{C}oncile: {R}ound-table conference improves reasoning via consensus among diverse {LLMs}.
\newblock \emph{arXiv e-prints, arXiv:2309.13007}, 2023.

\bibitem[Chu et~al.(2023)Chu, Chen, Chen, Yu, He, Wang, Peng, Liu, Qin, and Liu]{chu2023survey}
Z.~Chu, J.~Chen, Q.~Chen, W.~Yu, T.~He, H.~Wang, W.~Peng, M.~Liu, B.~Qin, and T.~Liu.
\newblock A survey of chain of thought reasoning: {A}dvances, frontiers and future.
\newblock \emph{arXiv preprint arxiv:2309.15402}, 2023.

\bibitem[Cobbe et~al.(2021)Cobbe, Kosaraju, Bavarian, Chen, Jun, Kaiser, Plappert, Tworek, Hilton, Nakano, Hesse, and Schulman]{cobbe2021gsm8k}
K.~Cobbe, V.~Kosaraju, M.~Bavarian, M.~Chen, H.~Jun, L.~Kaiser, M.~Plappert, J.~Tworek, J.~Hilton, R.~Nakano, C.~Hesse, and J.~Schulman.
\newblock Training verifiers to solve math word problems.
\newblock \emph{arXiv preprint arXiv:2110.14168}, 2021.

\bibitem[{Du} et~al.(2023){Du}, {Li}, {Torralba}, {Tenenbaum}, and {Mordatch}]{du2023debate}
Y.~{Du}, S.~{Li}, A.~{Torralba}, J.~B. {Tenenbaum}, and I.~{Mordatch}.
\newblock Improving factuality and reasoning in language models through multiagent debate.
\newblock \emph{arXiv e-prints, arXiv:2305.14325}, 2023.

\bibitem[Fu et~al.(2024)Fu, Huangfu, Yan, Ng, and Qiu]{fu2024hintbeforesolving}
J.~Fu, S.~Huangfu, H.~Yan, S.-K. Ng, and X.~Qiu.
\newblock Hint-before-solving prompting: {G}uiding {LLM}s to effectively utilize encoded knowledge.
\newblock \emph{arXiv preprint arXiv:2402.14310}, 2024.

\bibitem[{Fu} et~al.(2023){Fu}, {Peng}, {Sabharwal}, {Clark}, and {Khot}]{fu2023}
Y.~{Fu}, H.~{Peng}, A.~{Sabharwal}, P.~{Clark}, and T.~{Khot}.
\newblock Complexity-based prompting for multi-step reasoning.
\newblock In \emph{Proc. Int. Conf. Learn. Representations}, 2023.

\bibitem[Gou et~al.(2023)Gou, Shao, Gong, Shen, Yang, Duan, and Chen]{gou2023critic}
Z.~Gou, Z.~Shao, Y.~Gong, Y.~Shen, Y.~Yang, N.~Duan, and W.~Chen.
\newblock Critic: {L}arge language models can self-correct with tool-interactive critiquing.
\newblock \emph{arXiv preprint arXiv:2305.11738}, 2023.

\bibitem[{Grattafiori} et~al.(2024){Grattafiori}, {Dubey}, {Jauhri}, {Pandey}, {Kadian}, {Al-Dahle}, {Letman}, {Mathur}, {Schelten}, {Vaughan}, {Yang}, {Fan}, {Goyal}, {Hartshorn}, {Yang}, {Mitra}, {Sravankumar}, {Korenev}, {Hinsvark}, {Rao}, {Zhang}, {Rodriguez}, {Gregerson}, {Spataru}, {Roziere}, {Biron}, {Tang}, {Chern}, {Caucheteux}, {Nayak}, {Bi}, {Marra}, {McConnell}, {Keller}, {Touret}, {Wu}, {Wong}, {Canton Ferrer}, {Nikolaidis}, {Allonsius}, {Song}, {Pintz}, {Livshits}, {Wyatt}, {Esiobu}, {Choudhary}, {Mahajan}, {Garcia-Olano}, {Perino}, {Hupkes}, {Lakomkin}, {AlBadawy}, {Lobanova}, {Dinan}, {Smith}, {Radenovic}, {Guzm{\'a}n}, {Zhang}, {Synnaeve}, {Lee}, {Anderson}, {Thattai}, {Nail}, {Mialon}, {Pang}, {Cucurell}, {Nguyen}, {Korevaar}, {Xu}, {Touvron}, {Zarov}, {Arrieta Ibarra}, {Kloumann}, {Misra}, {Evtimov}, {Zhang}, {Copet}, {Lee}, {Geffert}, {Vranes}, {Park}, {Mahadeokar}, {Shah}, {van der Linde}, {Billock}, {Hong}, {Lee}, {Fu}, {Chi}, {Huang}, {Liu}, {Wang}, {Yu}, {Bitton}, {Spisak}, {Park},
  {Rocca}, {Johnstun}, {Saxe}, {Jia}, {Vasuden Alwala}, {Prasad}, {Upasani}, {Plawiak}, {Li}, {Heafield}, {Stone}, {El-Arini}, {Iyer}, {Malik}, {Chiu}, {Bhalla}, {Lakhotia}, {Rantala-Yeary}, {van der Maaten}, {Chen}, {Tan}, {Jenkins}, {Martin}, {Madaan}, {Malo}, {Blecher}, {Landzaat}, {de Oliveira}, {Muzzi}, {Pasupuleti}, {Singh}, {Paluri}, {Kardas}, {Tsimpoukelli}, {Oldham}, {Rita}, {Pavlova}, {Kambadur}, {Lewis}, {Si}, {Singh}, {Hassan}, {Goyal}, {Torabi}, {Bashlykov}, {Bogoychev}, {Chatterji}, {Zhang}, {Duchenne}, {{\c{C}}elebi}, {Alrassy}, {Zhang}, {Li}, {Vasic}, {Weng}, {Bhargava}, {Dubal}, {Krishnan}, {Singh Koura}, {Xu}, {He}, {Dong}, {Srinivasan}, {Ganapathy}, {Calderer}, {Silveira Cabral}, {Stojnic}, {Raileanu}, {Maheswari}, {Girdhar}, {Patel}, {Sauvestre}, {Polidoro}, {Sumbaly}, {Taylor}, {Silva}, {Hou}, {Wang}, {Hosseini}, {Chennabasappa}, {Singh}, {Bell}, {Kim}, {Edunov}, {Nie}, {Narang}, {Raparthy}, {Shen}, {Wan}, {Bhosale}, {Zhang}, {Vandenhende}, {Batra}, {Whitman}, {Sootla}, {Collot},
  {Gururangan}, {Borodinsky}, {Herman}, {Fowler}, {Sheasha}, {Georgiou}, {Scialom}, and {Speckbacher}]{llama3_2024}
A.~{Grattafiori}, A.~{Dubey}, A.~{Jauhri}, A.~{Pandey}, A.~{Kadian}, A.~{Al-Dahle}, A.~{Letman}, A.~{Mathur}, A.~{Schelten}, A.~{Vaughan}, A.~{Yang}, A.~{Fan}, A.~{Goyal}, A.~{Hartshorn}, A.~{Yang}, A.~{Mitra}, A.~{Sravankumar}, A.~{Korenev}, A.~{Hinsvark}, A.~{Rao}, A.~{Zhang}, A.~{Rodriguez}, A.~{Gregerson}, A.~{Spataru}, B.~{Roziere}, B.~{Biron}, B.~{Tang}, B.~{Chern}, C.~{Caucheteux}, C.~{Nayak}, C.~{Bi}, C.~{Marra}, C.~{McConnell}, C.~{Keller}, C.~{Touret}, C.~{Wu}, C.~{Wong}, C.~{Canton Ferrer}, C.~{Nikolaidis}, D.~{Allonsius}, D.~{Song}, D.~{Pintz}, D.~{Livshits}, D.~{Wyatt}, D.~{Esiobu}, D.~{Choudhary}, D.~{Mahajan}, D.~{Garcia-Olano}, D.~{Perino}, D.~{Hupkes}, E.~{Lakomkin}, E.~{AlBadawy}, E.~{Lobanova}, E.~{Dinan}, E.~M. {Smith}, F.~{Radenovic}, F.~{Guzm{\'a}n}, F.~{Zhang}, G.~{Synnaeve}, G.~{Lee}, G.~L. {Anderson}, G.~{Thattai}, G.~{Nail}, G.~{Mialon}, G.~{Pang}, G.~{Cucurell}, H.~{Nguyen}, H.~{Korevaar}, H.~{Xu}, H.~{Touvron}, I.~{Zarov}, I.~{Arrieta Ibarra}, I.~{Kloumann}, I.~{Misra},
  I.~{Evtimov}, J.~{Zhang}, J.~{Copet}, J.~{Lee}, J.~{Geffert}, J.~{Vranes}, J.~{Park}, J.~{Mahadeokar}, J.~{Shah}, J.~{van der Linde}, J.~{Billock}, J.~{Hong}, J.~{Lee}, J.~{Fu}, J.~{Chi}, J.~{Huang}, J.~{Liu}, J.~{Wang}, J.~{Yu}, J.~{Bitton}, J.~{Spisak}, J.~{Park}, J.~{Rocca}, J.~{Johnstun}, J.~{Saxe}, J.~{Jia}, K.~{Vasuden Alwala}, K.~{Prasad}, K.~{Upasani}, K.~{Plawiak}, K.~{Li}, K.~{Heafield}, K.~{Stone}, K.~{El-Arini}, K.~{Iyer}, K.~{Malik}, K.~{Chiu}, K.~{Bhalla}, K.~{Lakhotia}, L.~{Rantala-Yeary}, L.~{van der Maaten}, L.~{Chen}, L.~{Tan}, L.~{Jenkins}, L.~{Martin}, L.~{Madaan}, L.~{Malo}, L.~{Blecher}, L.~{Landzaat}, L.~{de Oliveira}, M.~{Muzzi}, M.~{Pasupuleti}, M.~{Singh}, M.~{Paluri}, M.~{Kardas}, M.~{Tsimpoukelli}, M.~{Oldham}, M.~{Rita}, M.~{Pavlova}, M.~{Kambadur}, M.~{Lewis}, M.~{Si}, M.~K. {Singh}, M.~{Hassan}, N.~{Goyal}, N.~{Torabi}, N.~{Bashlykov}, N.~{Bogoychev}, N.~{Chatterji}, N.~{Zhang}, O.~{Duchenne}, O.~{{\c{C}}elebi}, P.~{Alrassy}, P.~{Zhang}, P.~{Li}, P.~{Vasic}, P.~{Weng},
  P.~{Bhargava}, P.~{Dubal}, P.~{Krishnan}, P.~{Singh Koura}, P.~{Xu}, Q.~{He}, Q.~{Dong}, R.~{Srinivasan}, R.~{Ganapathy}, R.~{Calderer}, R.~{Silveira Cabral}, R.~{Stojnic}, R.~{Raileanu}, R.~{Maheswari}, R.~{Girdhar}, R.~{Patel}, R.~{Sauvestre}, R.~{Polidoro}, R.~{Sumbaly}, R.~{Taylor}, R.~{Silva}, R.~{Hou}, R.~{Wang}, S.~{Hosseini}, S.~{Chennabasappa}, S.~{Singh}, S.~{Bell}, S.~S. {Kim}, S.~{Edunov}, S.~{Nie}, S.~{Narang}, S.~{Raparthy}, S.~{Shen}, S.~{Wan}, S.~{Bhosale}, S.~{Zhang}, S.~{Vandenhende}, S.~{Batra}, S.~{Whitman}, S.~{Sootla}, S.~{Collot}, S.~{Gururangan}, S.~{Borodinsky}, T.~{Herman}, T.~{Fowler}, T.~{Sheasha}, T.~{Georgiou}, T.~{Scialom}, and T.~{Speckbacher}.
\newblock The {L}lama 3 herd of models.
\newblock \emph{arXiv e-prints, arXiv:2407.21783}, 2024.

\bibitem[Hendrycks et~al.(2021)Hendrycks, Burns, Kadavath, Arora, Basart, Tang, Song, and Steinhardt]{hendrycksmath2021}
D.~Hendrycks, C.~Burns, S.~Kadavath, A.~Arora, S.~Basart, E.~Tang, D.~Song, and J.~Steinhardt.
\newblock Measuring mathematical problem solving with the math dataset.
\newblock In \emph{Proc. Adv. Neural Inf. Process. Syst.}, 2021.

\bibitem[Hosseini et~al.(2014)Hosseini, Hajishirzi, Etzioni, and Kushman]{hosseini2014}
M.~J. Hosseini, H.~Hajishirzi, O.~Etzioni, and N.~Kushman.
\newblock Learning to solve arithmetic word problems with verb categorization.
\newblock In \emph{Proc. Conf. Empirical Methods in Natural Language Process.}, 2014.

\bibitem[Huang et~al.(2024)Huang, Chen, Mishra, Zheng, Yu, Song, and Zhou]{huang2023large}
J.~Huang, X.~Chen, S.~Mishra, H.~S. Zheng, A.~W. Yu, X.~Song, and D.~Zhou.
\newblock Large language models cannot self-correct reasoning yet.
\newblock In \emph{Proc. Int. Conf. Learn. Representations}, 2024.

\bibitem[Jiang et~al.(2024)Jiang, Shi, Yu, Liu, Zhang, Li, and Kwok]{jiang2024fobar}
W.~Jiang, H.~Shi, L.~Yu, Z.~Liu, Y.~Zhang, Z.~Li, and J.~T. Kwok.
\newblock Forward-backward reasoning in large language models for mathematical verification.
\newblock In \emph{Proc. Findings of Annual Meeting of the Assoc. Comput. Linguist.}, 2024.

\bibitem[Kojima et~al.(2022)Kojima, Gu, Reid, Matsuo, and Iwasawa]{kojima2022large}
T.~Kojima, S.~S. Gu, M.~Reid, Y.~Matsuo, and Y.~Iwasawa.
\newblock Large language models are zero-shot reasoners.
\newblock In \emph{Proc. Adv. Neural Inf. Process. Syst.}, pages 22199--22213, 2022.

\bibitem[Koncel-Kedziorski et~al.(2015)Koncel-Kedziorski, Hajishirzi, Sabharwal, Etzioni, and Ang]{koncel-kedziorski2015}
R.~Koncel-Kedziorski, H.~Hajishirzi, A.~Sabharwal, O.~Etzioni, and S.~D. Ang.
\newblock Parsing algebraic word problems into equations.
\newblock \emph{Trans. Assoc. Comput. Linguist.}, pages 585--597, 2015.

\bibitem[Lei et~al.(2024)Lei, Zhang, Zuo, Payani, and Ding]{lei2024macm}
B.~Lei, Y.~Zhang, S.~Zuo, A.~Payani, and C.~Ding.
\newblock {MACM}: Utilizing a multi-agent system for condition mining in solving complex mathematical problems.
\newblock In \emph{Proc. Adv. Neural Inf. Process. Syst.}, 2024.

\bibitem[Li et~al.(2023)Li, Wang, Guo, Song, Tan, Hassan, Menezes, Xiao, Bian, and Zhu]{li2023deliberate}
B.~Li, R.~Wang, J.~Guo, K.~Song, X.~Tan, H.~Hassan, A.~Menezes, T.~Xiao, J.~Bian, and J.~Zhu.
\newblock Deliberate then generate: Enhanced prompting framework for text generation.
\newblock \emph{arXiv preprint arXiv:2305.19835}, 2023.

\bibitem[Ling et~al.(2017)Ling, Yogatama, Dyer, and Blunsom]{ling2017}
W.~Ling, D.~Yogatama, C.~Dyer, and P.~Blunsom.
\newblock Program induction by rationale generation: Learning to solve and explain algebraic word problems.
\newblock In \emph{Proc. Conf. Empirical Methods in Natural Language Process.}, 2017.

\bibitem[Madaan et~al.(2023)Madaan, Tandon, Gupta, Hallinan, Gao, Wiegreffe, Alon, Dziri, Prabhumoye, Yang, et~al.]{madaan2024self}
A.~Madaan, N.~Tandon, P.~Gupta, S.~Hallinan, L.~Gao, S.~Wiegreffe, U.~Alon, N.~Dziri, S.~Prabhumoye, Y.~Yang, et~al.
\newblock Self-refine: {I}terative refinement with self-feedback.
\newblock In \emph{Proc. Adv. Neural Inf. Process. Syst.}, 2023.

\bibitem[OpenAI et~al.(2024)OpenAI, Achiam, Adler, Agarwal, Ahmad, Akkaya, Aleman, Almeida, Altenschmidt, Altman, Anadkat, Avila, Babuschkin, Balaji, Balcom, Baltescu, Bao, Bavarian, Belgum, Bello, Berdine, Bernadett-Shapiro, Berner, Bogdonoff, Boiko, Boyd, Brakman, Brockman, Brooks, Brundage, Button, Cai, Campbell, Cann, Carey, Carlson, Carmichael, Chan, Chang, Chantzis, Chen, Chen, Chen, Chen, Chen, Chess, Cho, Chu, Chung, Cummings, Currier, Dai, Decareaux, Degry, Deutsch, Deville, Dhar, Dohan, Dowling, Dunning, Ecoffet, Eleti, Eloundou, Farhi, Fedus, Felix, Fishman, Forte, Fulford, Gao, Georges, Gibson, Goel, Gogineni, Goh, Gontijo-Lopes, Gordon, Grafstein, Gray, Greene, Gross, Gu, Guo, Hallacy, Han, Harris, He, Heaton, Heidecke, Hesse, Hickey, Hickey, Hoeschele, Houghton, Hsu, Hu, Hu, Huizinga, Jain, Jain, Jang, Jiang, Jiang, Jin, Jin, Jomoto, Jonn, Jun, Kaftan, Łukasz Kaiser, Kamali, Kanitscheider, Keskar, Khan, Kilpatrick, Kim, Kim, Kim, Kirchner, Kiros, Knight, Kokotajlo, Łukasz Kondraciuk, Kondrich,
  Konstantinidis, Kosic, Krueger, Kuo, Lampe, Lan, Lee, Leike, Leung, Levy, Li, Lim, Lin, Lin, Litwin, Lopez, Lowe, Lue, Makanju, Malfacini, Manning, Markov, Markovski, Martin, Mayer, Mayne, McGrew, McKinney, McLeavey, McMillan, McNeil, Medina, Mehta, Menick, Metz, Mishchenko, Mishkin, Monaco, Morikawa, Mossing, Mu, Murati, Murk, Mély, Nair, Nakano, Nayak, Neelakantan, Ngo, Noh, Ouyang, O'Keefe, Pachocki, Paino, Palermo, Pantuliano, Parascandolo, Parish, Parparita, Passos, Pavlov, Peng, Perelman, de~Avila Belbute~Peres, Petrov, de~Oliveira~Pinto, Michael, Pokorny, Pokrass, Pong, Powell, Power, Power, Proehl, Puri, Radford, Rae, Ramesh, Raymond, Real, Rimbach, Ross, Rotsted, Roussez, Ryder, Saltarelli, Sanders, Santurkar, Sastry, Schmidt, Schnurr, Schulman, Selsam, Sheppard, Sherbakov, Shieh, Shoker, Shyam, Sidor, Sigler, Simens, Sitkin, Slama, Sohl, Sokolowsky, Song, Staudacher, Such, Summers, Sutskever, Tang, Tezak, Thompson, Tillet, Tootoonchian, Tseng, Tuggle, Turley, Tworek, Uribe, Vallone, Vijayvergiya,
  Voss, Wainwright, Wang, Wang, Wang, Ward, Wei, Weinmann, Welihinda, Welinder, Weng, Weng, Wiethoff, Willner, Winter, Wolrich, Wong, Workman, Wu, Wu, Wu, Xiao, Xu, Yoo, Yu, Yuan, Zaremba, Zellers, Zhang, Zhang, Zhao, Zheng, Zhuang, Zhuk, and Zoph]{openai2024gpt4}
OpenAI, J.~Achiam, S.~Adler, S.~Agarwal, L.~Ahmad, I.~Akkaya, F.~L. Aleman, D.~Almeida, J.~Altenschmidt, S.~Altman, S.~Anadkat, R.~Avila, I.~Babuschkin, S.~Balaji, V.~Balcom, P.~Baltescu, H.~Bao, M.~Bavarian, J.~Belgum, I.~Bello, J.~Berdine, G.~Bernadett-Shapiro, C.~Berner, L.~Bogdonoff, O.~Boiko, M.~Boyd, A.-L. Brakman, G.~Brockman, T.~Brooks, M.~Brundage, K.~Button, T.~Cai, R.~Campbell, A.~Cann, B.~Carey, C.~Carlson, R.~Carmichael, B.~Chan, C.~Chang, F.~Chantzis, D.~Chen, S.~Chen, R.~Chen, J.~Chen, M.~Chen, B.~Chess, C.~Cho, C.~Chu, H.~W. Chung, D.~Cummings, J.~Currier, Y.~Dai, C.~Decareaux, T.~Degry, N.~Deutsch, D.~Deville, A.~Dhar, D.~Dohan, S.~Dowling, S.~Dunning, A.~Ecoffet, A.~Eleti, T.~Eloundou, D.~Farhi, L.~Fedus, N.~Felix, S.~P. Fishman, J.~Forte, I.~Fulford, L.~Gao, E.~Georges, C.~Gibson, V.~Goel, T.~Gogineni, G.~Goh, R.~Gontijo-Lopes, J.~Gordon, M.~Grafstein, S.~Gray, R.~Greene, J.~Gross, S.~S. Gu, Y.~Guo, C.~Hallacy, J.~Han, J.~Harris, Y.~He, M.~Heaton, J.~Heidecke, C.~Hesse, A.~Hickey,
  W.~Hickey, P.~Hoeschele, B.~Houghton, K.~Hsu, S.~Hu, X.~Hu, J.~Huizinga, S.~Jain, S.~Jain, J.~Jang, A.~Jiang, R.~Jiang, H.~Jin, D.~Jin, S.~Jomoto, B.~Jonn, H.~Jun, T.~Kaftan, Łukasz Kaiser, A.~Kamali, I.~Kanitscheider, N.~S. Keskar, T.~Khan, L.~Kilpatrick, J.~W. Kim, C.~Kim, Y.~Kim, J.~H. Kirchner, J.~Kiros, M.~Knight, D.~Kokotajlo, Łukasz Kondraciuk, A.~Kondrich, A.~Konstantinidis, K.~Kosic, G.~Krueger, V.~Kuo, M.~Lampe, I.~Lan, T.~Lee, J.~Leike, J.~Leung, D.~Levy, C.~M. Li, R.~Lim, M.~Lin, S.~Lin, M.~Litwin, T.~Lopez, R.~Lowe, P.~Lue, A.~Makanju, K.~Malfacini, S.~Manning, T.~Markov, Y.~Markovski, B.~Martin, K.~Mayer, A.~Mayne, B.~McGrew, S.~M. McKinney, C.~McLeavey, P.~McMillan, J.~McNeil, D.~Medina, A.~Mehta, J.~Menick, L.~Metz, A.~Mishchenko, P.~Mishkin, V.~Monaco, E.~Morikawa, D.~Mossing, T.~Mu, M.~Murati, O.~Murk, D.~Mély, A.~Nair, R.~Nakano, R.~Nayak, A.~Neelakantan, R.~Ngo, H.~Noh, L.~Ouyang, C.~O'Keefe, J.~Pachocki, A.~Paino, J.~Palermo, A.~Pantuliano, G.~Parascandolo, J.~Parish, E.~Parparita,
  A.~Passos, M.~Pavlov, A.~Peng, A.~Perelman, F.~de~Avila Belbute~Peres, M.~Petrov, H.~P. de~Oliveira~Pinto, Michael, Pokorny, M.~Pokrass, V.~H. Pong, T.~Powell, A.~Power, B.~Power, E.~Proehl, R.~Puri, A.~Radford, J.~Rae, A.~Ramesh, C.~Raymond, F.~Real, K.~Rimbach, C.~Ross, B.~Rotsted, H.~Roussez, N.~Ryder, M.~Saltarelli, T.~Sanders, S.~Santurkar, G.~Sastry, H.~Schmidt, D.~Schnurr, J.~Schulman, D.~Selsam, K.~Sheppard, T.~Sherbakov, J.~Shieh, S.~Shoker, P.~Shyam, S.~Sidor, E.~Sigler, M.~Simens, J.~Sitkin, K.~Slama, I.~Sohl, B.~Sokolowsky, Y.~Song, N.~Staudacher, F.~P. Such, N.~Summers, I.~Sutskever, J.~Tang, N.~Tezak, M.~B. Thompson, P.~Tillet, A.~Tootoonchian, E.~Tseng, P.~Tuggle, N.~Turley, J.~Tworek, J.~F.~C. Uribe, A.~Vallone, A.~Vijayvergiya, C.~Voss, C.~Wainwright, J.~J. Wang, A.~Wang, B.~Wang, J.~Ward, J.~Wei, C.~Weinmann, A.~Welihinda, P.~Welinder, J.~Weng, L.~Weng, M.~Wiethoff, D.~Willner, C.~Winter, S.~Wolrich, H.~Wong, L.~Workman, S.~Wu, J.~Wu, M.~Wu, K.~Xiao, T.~Xu, S.~Yoo, K.~Yu, Q.~Yuan,
  W.~Zaremba, R.~Zellers, C.~Zhang, M.~Zhang, S.~Zhao, T.~Zheng, J.~Zhuang, W.~Zhuk, and B.~Zoph.
\newblock {GPT}-4 technical report.
\newblock \emph{arXiv preprint arXiv:2303.08774}, 2024.

\bibitem[Park et~al.(2023)Park, O'Brien, Cai, Morris, Liang, and Bernstein]{park2023generative}
J.~S. Park, J.~O'Brien, C.~J. Cai, M.~R. Morris, P.~Liang, and M.~S. Bernstein.
\newblock Generative agents: {I}nteractive simulacra of human behavior.
\newblock In \emph{Proc. ACM Symp. User Interface Software and Technology}, 2023.

\bibitem[Patel et~al.(2021)Patel, Bhattamishra, and Goyal]{patel2021}
A.~Patel, S.~Bhattamishra, and N.~Goyal.
\newblock Are {NLP} models really able to solve simple math word problems?
\newblock In \emph{Proc. Conf. North Amer. Chapter of the Associ. Comput. Linguist.: Human Language Technologies}, 2021.

\bibitem[Paul et~al.(2023)Paul, Ismayilzada, Peyrard, Borges, Bosselut, West, and Faltings]{paul2023refiner}
D.~Paul, M.~Ismayilzada, M.~Peyrard, B.~Borges, A.~Bosselut, R.~West, and B.~Faltings.
\newblock Refiner: {R}easoning feedback on intermediate representations.
\newblock \emph{arXiv preprint arXiv:2304.01904}, 2023.

\bibitem[Roy and Roth(2015)]{roy2015}
S.~Roy and D.~Roth.
\newblock Solving general arithmetic word problems.
\newblock In \emph{Proc. Conf. Empirical Methods in Natural Language Process.}, 2015.

\bibitem[Saparov and He(2022)]{saparov2022language}
A.~Saparov and H.~He.
\newblock Language models are greedy reasoners: A systematic formal analysis of chain-of-thought.
\newblock \emph{arXiv preprint arXiv:2210.01240}, 2022.

\bibitem[Shinn et~al.(2023)Shinn, Cassano, Gopinath, Narasimhan, and Yao]{shinn2024reflexion}
N.~Shinn, F.~Cassano, A.~Gopinath, K.~Narasimhan, and S.~Yao.
\newblock Reflexion: {L}anguage agents with verbal reinforcement learning.
\newblock In \emph{Proc. Adv. Neural Inf. Process. Syst.}, 2023.

\bibitem[Srivastava et~al.(2023)Srivastava, Rastogi, Rao, Shoeb, Abid, Fisch, Brown, Santoro, Gupta, Garriga-Alonso, Kluska, Lewkowycz, Agarwal, Power, Ray, Warstadt, Kocurek, Safaya, Tazarv, Xiang, Parrish, Nie, Hussain, Askell, Dsouza, Slone, Rahane, Iyer, Andreassen, Madotto, Santilli, Stuhlm{\"u}ller, Dai, La, Lampinen, Zou, Jiang, Chen, Vuong, Gupta, Gottardi, Norelli, Venkatesh, Gholamidavoodi, Tabassum, Menezes, Kirubarajan, Mullokandov, Sabharwal, Herrick, Efrat, Erdem, Karaka{\c{s}}, Roberts, Loe, Zoph, Bojanowski, {\"O}zyurt, Hedayatnia, Neyshabur, Inden, Stein, Ekmekci, Lin, Howald, Orinion, Diao, Dour, Stinson, Argueta, Ferri, Singh, Rathkopf, Meng, Baral, Wu, Callison-Burch, Waites, Voigt, Manning, Potts, Ramirez, Rivera, Siro, Raffel, Ashcraft, Garbacea, Sileo, Garrette, Hendrycks, Kilman, Roth, Freeman, Khashabi, Levy, Gonz{\'a}lez, Perszyk, Hernandez, Chen, Ippolito, Gilboa, Dohan, Drakard, Jurgens, Datta, Ganguli, Emelin, Kleyko, Yuret, Chen, Tam, Hupkes, Misra, Buzan, Mollo, Yang, Lee,
  Schrader, Shutova, Cubuk, Segal, Hagerman, Barnes, Donoway, Pavlick, Rodol{\`a}, Lam, Chu, Tang, Erdem, Chang, Chi, Dyer, Jerzak, Kim, Manyasi, Zheltonozhskii, Xia, Siar, Mart{\'\i}nez-Plumed, Happ{\'e}, Chollet, Rong, Mishra, Winata, de~Melo, Kruszewski, Parascandolo, Mariani, Wang, Jaimovitch-Lopez, Betz, Gur-Ari, Galijasevic, Kim, Rashkin, Hajishirzi, Mehta, Bogar, Shevlin, Schuetze, Yakura, Zhang, Wong, Ng, Noble, Jumelet, Geissinger, Kernion, Hilton, Lee, Fisac, Simon, Koppel, Zheng, Zou, Kocon, Thompson, Wingfield, Kaplan, Radom, Sohl-Dickstein, Phang, Wei, Yosinski, Novikova, Bosscher, Marsh, Kim, Taal, Engel, Alabi, Xu, Song, Tang, Waweru, Burden, Miller, Balis, Batchelder, Berant, Frohberg, Rozen, Hernandez-Orallo, Boudeman, Guerr, Jones, Tenenbaum, Rule, Chua, Kanclerz, Livescu, Krauth, Gopalakrishnan, Ignatyeva, Markert, Dhole, Gimpel, Omondi, Mathewson, Chiafullo, Shkaruta, Shridhar, McDonell, Richardson, Reynolds, Gao, Zhang, Dugan, Qin, Contreras-Ochando, Morency, Moschella, Lam, Noble,
  Schmidt, He, Oliveros-Col{\'o}n, Metz, Senel, Bosma, Sap, Hoeve, Farooqi, Faruqui, Mazeika, Baturan, Marelli, Maru, Ramirez-Quintana, Tolkiehn, Giulianelli, Lewis, Potthast, Leavitt, Hagen, Schubert, Baitemirova, Arnaud, McElrath, Yee, Cohen, Gu, Ivanitskiy, Starritt, Strube, Sw{e}drowski, Bevilacqua, Yasunaga, Kale, Cain, Xu, Suzgun, Walker, Tiwari, Bansal, Aminnaseri, Geva, Gheini, T, Peng, Chi, Lee, Krakover, Cameron, Roberts, Doiron, Martinez, Nangia, Deckers, Muennighoff, Keskar, Iyer, Constant, Fiedel, Wen, Zhang, Agha, Elbaghdadi, Levy, Evans, Casares, Doshi, Fung, Liang, Vicol, Alipoormolabashi, Liao, Liang, Chang, Eckersley, Htut, Hwang, Mi{\l}kowski, Patil, Pezeshkpour, Oli, Mei, Lyu, Chen, Banjade, Rudolph, Gabriel, Habacker, Risco, Milli{\`e}re, Garg, Barnes, Saurous, Arakawa, Raymaekers, Frank, Sikand, Novak, Sitelew, Bras, Liu, Jacobs, Zhang, Salakhutdinov, Chi, Lee, Stovall, Teehan, Yang, Singh, Mohammad, Anand, Dillavou, Shleifer, Wiseman, Gruetter, Bowman, Schoenholz, Han, Kwatra, Rous,
  Ghazarian, Ghosh, Casey, Bischoff, Gehrmann, Schuster, Sadeghi, Hamdan, Zhou, Srivastava, Shi, Singh, Asaadi, Gu, Pachchigar, Toshniwal, Upadhyay, Debnath, Shakeri, Thormeyer, Melzi, Reddy, Makini, Lee, Torene, Hatwar, Dehaene, Divic, Ermon, Biderman, Lin, Prasad, Piantadosi, Shieber, Misherghi, Kiritchenko, Mishra, Linzen, Schuster, Li, Yu, Ali, Hashimoto, Wu, Desbordes, Rothschild, Phan, Wang, Nkinyili, Schick, Kornev, Tunduny, Gerstenberg, Chang, Neeraj, Khot, Shultz, Shaham, Misra, Demberg, Nyamai, Raunak, Ramasesh, vinay~uday prabhu, Padmakumar, Srikumar, Fedus, Saunders, Zhang, Vossen, Ren, Tong, Zhao, Wu, Shen, Yaghoobzadeh, Lakretz, Song, Bahri, Choi, Yang, Hao, Chen, Belinkov, Hou, Hou, Bai, Seid, Zhao, Wang, Wang, Wang, and Wu]{srivastava2023beyond}
A.~Srivastava, A.~Rastogi, A.~Rao, A.~A.~M. Shoeb, A.~Abid, A.~Fisch, A.~R. Brown, A.~Santoro, A.~Gupta, A.~Garriga-Alonso, A.~Kluska, A.~Lewkowycz, A.~Agarwal, A.~Power, A.~Ray, A.~Warstadt, A.~W. Kocurek, A.~Safaya, A.~Tazarv, A.~Xiang, A.~Parrish, A.~Nie, A.~Hussain, A.~Askell, A.~Dsouza, A.~Slone, A.~Rahane, A.~S. Iyer, A.~J. Andreassen, A.~Madotto, A.~Santilli, A.~Stuhlm{\"u}ller, A.~M. Dai, A.~La, A.~K. Lampinen, A.~Zou, A.~Jiang, A.~Chen, A.~Vuong, A.~Gupta, A.~Gottardi, A.~Norelli, A.~Venkatesh, A.~Gholamidavoodi, A.~Tabassum, A.~Menezes, A.~Kirubarajan, A.~Mullokandov, A.~Sabharwal, A.~Herrick, A.~Efrat, A.~Erdem, A.~Karaka{\c{s}}, B.~R. Roberts, B.~S. Loe, B.~Zoph, B.~Bojanowski, B.~{\"O}zyurt, B.~Hedayatnia, B.~Neyshabur, B.~Inden, B.~Stein, B.~Ekmekci, B.~Y. Lin, B.~Howald, B.~Orinion, C.~Diao, C.~Dour, C.~Stinson, C.~Argueta, C.~Ferri, C.~Singh, C.~Rathkopf, C.~Meng, C.~Baral, C.~Wu, C.~Callison-Burch, C.~Waites, C.~Voigt, C.~D. Manning, C.~Potts, C.~Ramirez, C.~E. Rivera, C.~Siro, C.~Raffel,
  C.~Ashcraft, C.~Garbacea, D.~Sileo, D.~Garrette, D.~Hendrycks, D.~Kilman, D.~Roth, C.~D. Freeman, D.~Khashabi, D.~Levy, D.~M. Gonz{\'a}lez, D.~Perszyk, D.~Hernandez, D.~Chen, D.~Ippolito, D.~Gilboa, D.~Dohan, D.~Drakard, D.~Jurgens, D.~Datta, D.~Ganguli, D.~Emelin, D.~Kleyko, D.~Yuret, D.~Chen, D.~Tam, D.~Hupkes, D.~Misra, D.~Buzan, D.~C. Mollo, D.~Yang, D.-H. Lee, D.~Schrader, E.~Shutova, E.~D. Cubuk, E.~Segal, E.~Hagerman, E.~Barnes, E.~Donoway, E.~Pavlick, E.~Rodol{\`a}, E.~Lam, E.~Chu, E.~Tang, E.~Erdem, E.~Chang, E.~A. Chi, E.~Dyer, E.~Jerzak, E.~Kim, E.~E. Manyasi, E.~Zheltonozhskii, F.~Xia, F.~Siar, F.~Mart{\'\i}nez-Plumed, F.~Happ{\'e}, F.~Chollet, F.~Rong, G.~Mishra, G.~I. Winata, G.~de~Melo, G.~Kruszewski, G.~Parascandolo, G.~Mariani, G.~X. Wang, G.~Jaimovitch-Lopez, G.~Betz, G.~Gur-Ari, H.~Galijasevic, H.~Kim, H.~Rashkin, H.~Hajishirzi, H.~Mehta, H.~Bogar, H.~F.~A. Shevlin, H.~Schuetze, H.~Yakura, H.~Zhang, H.~M. Wong, I.~Ng, I.~Noble, J.~Jumelet, J.~Geissinger, J.~Kernion, J.~Hilton, J.~Lee,
  J.~F. Fisac, J.~B. Simon, J.~Koppel, J.~Zheng, J.~Zou, J.~Kocon, J.~Thompson, J.~Wingfield, J.~Kaplan, J.~Radom, J.~Sohl-Dickstein, J.~Phang, J.~Wei, J.~Yosinski, J.~Novikova, J.~Bosscher, J.~Marsh, J.~Kim, J.~Taal, J.~Engel, J.~Alabi, J.~Xu, J.~Song, J.~Tang, J.~Waweru, J.~Burden, J.~Miller, J.~U. Balis, J.~Batchelder, J.~Berant, J.~Frohberg, J.~Rozen, J.~Hernandez-Orallo, J.~Boudeman, J.~Guerr, J.~Jones, J.~B. Tenenbaum, J.~S. Rule, J.~Chua, K.~Kanclerz, K.~Livescu, K.~Krauth, K.~Gopalakrishnan, K.~Ignatyeva, K.~Markert, K.~Dhole, K.~Gimpel, K.~Omondi, K.~W. Mathewson, K.~Chiafullo, K.~Shkaruta, K.~Shridhar, K.~McDonell, K.~Richardson, L.~Reynolds, L.~Gao, L.~Zhang, L.~Dugan, L.~Qin, L.~Contreras-Ochando, L.-P. Morency, L.~Moschella, L.~Lam, L.~Noble, L.~Schmidt, L.~He, L.~Oliveros-Col{\'o}n, L.~Metz, L.~K. Senel, M.~Bosma, M.~Sap, M.~T. Hoeve, M.~Farooqi, M.~Faruqui, M.~Mazeika, M.~Baturan, M.~Marelli, M.~Maru, M.~J. Ramirez-Quintana, M.~Tolkiehn, M.~Giulianelli, M.~Lewis, M.~Potthast, M.~L. Leavitt,
  M.~Hagen, M.~Schubert, M.~O. Baitemirova, M.~Arnaud, M.~McElrath, M.~A. Yee, M.~Cohen, M.~Gu, M.~Ivanitskiy, M.~Starritt, M.~Strube, M.~Sw{e}drowski, M.~Bevilacqua, M.~Yasunaga, M.~Kale, M.~Cain, M.~Xu, M.~Suzgun, M.~Walker, M.~Tiwari, M.~Bansal, M.~Aminnaseri, M.~Geva, M.~Gheini, M.~V. T, N.~Peng, N.~A. Chi, N.~Lee, N.~G.-A. Krakover, N.~Cameron, N.~Roberts, N.~Doiron, N.~Martinez, N.~Nangia, N.~Deckers, N.~Muennighoff, N.~S. Keskar, N.~S. Iyer, N.~Constant, N.~Fiedel, N.~Wen, O.~Zhang, O.~Agha, O.~Elbaghdadi, O.~Levy, O.~Evans, P.~A.~M. Casares, P.~Doshi, P.~Fung, P.~P. Liang, P.~Vicol, P.~Alipoormolabashi, P.~Liao, P.~Liang, P.~W. Chang, P.~Eckersley, P.~M. Htut, P.~Hwang, P.~Mi{\l}kowski, P.~Patil, P.~Pezeshkpour, P.~Oli, Q.~Mei, Q.~Lyu, Q.~Chen, R.~Banjade, R.~E. Rudolph, R.~Gabriel, R.~Habacker, R.~Risco, R.~Milli{\`e}re, R.~Garg, R.~Barnes, R.~A. Saurous, R.~Arakawa, R.~Raymaekers, R.~Frank, R.~Sikand, R.~Novak, R.~Sitelew, R.~L. Bras, R.~Liu, R.~Jacobs, R.~Zhang, R.~Salakhutdinov, R.~A. Chi, S.~R.
  Lee, R.~Stovall, R.~Teehan, R.~Yang, S.~Singh, S.~M. Mohammad, S.~Anand, S.~Dillavou, S.~Shleifer, S.~Wiseman, S.~Gruetter, S.~R. Bowman, S.~S. Schoenholz, S.~Han, S.~Kwatra, S.~A. Rous, S.~Ghazarian, S.~Ghosh, S.~Casey, S.~Bischoff, S.~Gehrmann, S.~Schuster, S.~Sadeghi, S.~Hamdan, S.~Zhou, S.~Srivastava, S.~Shi, S.~Singh, S.~Asaadi, S.~S. Gu, S.~Pachchigar, S.~Toshniwal, S.~Upadhyay, S.~S. Debnath, S.~Shakeri, S.~Thormeyer, S.~Melzi, S.~Reddy, S.~P. Makini, S.-H. Lee, S.~Torene, S.~Hatwar, S.~Dehaene, S.~Divic, S.~Ermon, S.~Biderman, S.~Lin, S.~Prasad, S.~Piantadosi, S.~Shieber, S.~Misherghi, S.~Kiritchenko, S.~Mishra, T.~Linzen, T.~Schuster, T.~Li, T.~Yu, T.~Ali, T.~Hashimoto, T.-L. Wu, T.~Desbordes, T.~Rothschild, T.~Phan, T.~Wang, T.~Nkinyili, T.~Schick, T.~Kornev, T.~Tunduny, T.~Gerstenberg, T.~Chang, T.~Neeraj, T.~Khot, T.~Shultz, U.~Shaham, V.~Misra, V.~Demberg, V.~Nyamai, V.~Raunak, V.~V. Ramasesh, vinay~uday prabhu, V.~Padmakumar, V.~Srikumar, W.~Fedus, W.~Saunders, W.~Zhang, W.~Vossen, X.~Ren,
  X.~Tong, X.~Zhao, X.~Wu, X.~Shen, Y.~Yaghoobzadeh, Y.~Lakretz, Y.~Song, Y.~Bahri, Y.~Choi, Y.~Yang, S.~Hao, Y.~Chen, Y.~Belinkov, Y.~Hou, Y.~Hou, Y.~Bai, Z.~Seid, Z.~Zhao, Z.~Wang, Z.~J. Wang, Z.~Wang, and Z.~Wu.
\newblock Beyond the imitation game: Quantifying and extrapolating the capabilities of language models.
\newblock \emph{Transactions on Machine Learning Research}, 2023.
\newblock ISSN 2835-8856.

\bibitem[Suzgun et~al.(2023)Suzgun, Scales, Schärli, Gehrmann, Tay, Chung, Chowdhery, Le, Chi, Zhou, and Wei]{suzgun2023bbh}
M.~Suzgun, N.~Scales, N.~Schärli, S.~Gehrmann, Y.~Tay, H.~W. Chung, A.~Chowdhery, Q.~V. Le, E.~H. Chi, D.~Zhou, and J.~Wei.
\newblock Challenging big-bench tasks and whether chain-of-thought can solve them.
\newblock In \emph{Proc. ACL Findings}, 2023.

\bibitem[Tyen et~al.(2023)Tyen, Mansoor, Chen, Mak, and C{\u{a}}rbune]{tyen2023llms}
G.~Tyen, H.~Mansoor, P.~Chen, T.~Mak, and V.~C{\u{a}}rbune.
\newblock {LLM}s cannot find reasoning errors, but can correct them!
\newblock \emph{arXiv preprint arXiv:2311.08516}, 2023.

\bibitem[Wang et~al.(2023)Wang, Wei, Schuurmans, Le, Chi, Narang, Chowdhery, and Zhou]{wang2023selfconsistency}
X.~Wang, J.~Wei, D.~Schuurmans, Q.~V. Le, E.~H. Chi, S.~Narang, A.~Chowdhery, and D.~Zhou.
\newblock Self-consistency improves chain of thought reasoning in language models.
\newblock In \emph{Proc. Int. Conf. Learn. Representations}, 2023.

\bibitem[Wei et~al.(2022{\natexlab{a}})Wei, Tay, Bommasani, Raffel, Zoph, Borgeaud, Yogatama, Bosma, Zhou, Metzler, et~al.]{wei2022emergent}
J.~Wei, Y.~Tay, R.~Bommasani, C.~Raffel, B.~Zoph, S.~Borgeaud, D.~Yogatama, M.~Bosma, D.~Zhou, D.~Metzler, et~al.
\newblock Emergent abilities of large language models.
\newblock \emph{arXiv preprint arXiv:2206.07682}, 2022{\natexlab{a}}.

\bibitem[Wei et~al.(2022{\natexlab{b}})Wei, Wang, Schuurmans, Bosma, Xia, Chi, Le, Zhou, et~al.]{wei2022chain}
J.~Wei, X.~Wang, D.~Schuurmans, M.~Bosma, F.~Xia, E.~Chi, Q.~V. Le, D.~Zhou, et~al.
\newblock Chain-of-thought prompting elicits reasoning in large language models.
\newblock In \emph{Proc. Adv. Neural Inf. Process. Syst.}, 2022{\natexlab{b}}.

\bibitem[Weng et~al.(2023)Weng, Zhu, Xia, Li, He, Liu, Sun, Liu, and Zhao]{weng2023self-verify}
Y.~Weng, M.~Zhu, F.~Xia, B.~Li, S.~He, S.~Liu, B.~Sun, K.~Liu, and J.~Zhao.
\newblock Large language models are better reasoners with self-verification.
\newblock In \emph{Conf. Empirical Methods in Natural Language Process.}, 2023.

\bibitem[Wu et~al.(2024)Wu, Jiang, and Shen]{wu2024get}
Z.~Wu, M.~Jiang, and C.~Shen.
\newblock Get an {A} in math: {P}rogressive rectification prompting.
\newblock In \emph{Proc. AAAI Conf. Artif. Intell.}, 2024.

\bibitem[{Yao} et~al.(2023){Yao}, {Yu}, {Zhao}, {Shafran}, {Griffiths}, {Cao}, and {Narasimhan}]{yao2023}
S.~{Yao}, D.~{Yu}, J.~{Zhao}, I.~{Shafran}, T.~L. {Griffiths}, Y.~{Cao}, and K.~{Narasimhan}.
\newblock Tree of thoughts: {D}eliberate problem solving with large language models.
\newblock \emph{arXiv e-prints, arXiv:2305.10601}, 2023.

\bibitem[Yao et~al.(2023)Yao, Heinecke, Niebles, Liu, Feng, Xue, Murthy, Chen, Zhang, Arpit, et~al.]{yao2023retroformer}
W.~Yao, S.~Heinecke, J.~C. Niebles, Z.~Liu, Y.~Feng, L.~Xue, R.~Murthy, Z.~Chen, J.~Zhang, D.~Arpit, et~al.
\newblock Retroformer: {R}etrospective large language agents with policy gradient optimization.
\newblock \emph{arXiv preprint arXiv:2308.02151}, 2023.

\bibitem[{Zelikman} et~al.(2022){Zelikman}, {Wu}, {Mu}, and {Goodman}]{zelikman2022}
E.~{Zelikman}, Y.~{Wu}, J.~{Mu}, and N.~D. {Goodman}.
\newblock {STaR: Bootstrapping Reasoning With Reasoning}.
\newblock \emph{arXiv e-prints, arXiv:2203.14465}, 2022.

\bibitem[Zhang et~al.(2023{\natexlab{a}})Zhang, Cai, Zhang, Zhang, Mao, and Wu]{zhang2023self}
H.~Zhang, M.~Cai, X.~Zhang, C.~J. Zhang, R.~Mao, and K.~Wu.
\newblock Self-convinced prompting: {F}ew-shot question answering with repeated introspection.
\newblock \emph{arXiv preprint arXiv:2310.05035}, 2023{\natexlab{a}}.

\bibitem[Zhang et~al.(2023{\natexlab{b}})Zhang, Zhang, Li, and Smola]{zhang2023automatic}
Z.~Zhang, A.~Zhang, M.~Li, and A.~Smola.
\newblock Automatic chain of thought prompting in large language models.
\newblock In \emph{Proc. Int. Conf. Learn. Representations}, 2023{\natexlab{b}}.

\bibitem[Zheng et~al.(2023)Zheng, Liu, Xie, Li, and Li]{zheng2023progressive}
C.~Zheng, Z.~Liu, E.~Xie, Z.~Li, and Y.~Li.
\newblock Progressive-hint prompting improves reasoning in large language models.
\newblock \emph{arXiv preprint arXiv:2304.09797}, 2023.

\end{thebibliography}

\appendix
\section{Pseudocode of RAD}
\label{sec:alg}
\begin{algorithm}[ht]
\caption{Refined Answer Distribution (RAD)}
\label{alg:hm}
\begin{algorithmic}[1]
\STATE {\bfseries Input:} task $x$
\STATE {\bfseries Hyperparameters:} sampling budget $B_{max}>0$, number of iterations $R>1$ and $\{B_r>0\}_{r=1}^R$ such that $\sum_{r=1}^R B_r= B_{max}$
\STATE {\bfseries Output:} answer $\hat{y}$, approximations of $\{p_{r}(\tilde{y}|x)\}_{r=1}^R$
\vspace{0.1cm}
\FOR{$r=0:R-1$}
\IF{$r=0$}
\STATE Sample $B_1$ answers from  $p_1(\cdot|x)$.
\STATE Approximate $p_{1}(\tilde{y}|x)$ using eq.~\ref{eq:hm-mc-r}.
\ELSE
\FOR{$m=1:M$}
\STATE Sample $\lfloor \frac{B_{r{+}1}}{M} \rfloor$ answers from $ p(\cdot|x, \mathrm{Refine}(y^{m}))$ in order to form a Monte carlo approximation, as shown in eq.~\ref{eq:php-mc}.
\ENDFOR
\STATE Approximate $p_{r{+}1}(\tilde{y}|x)$ using eqs.~\ref{eq:hm-mc-r-plus-one} and~\ref{eq:hm-mc-one-answer}.
\ENDIF
\STATE Find the mode of the approximated $p_{R}(\tilde{y}|x)$ and assign it to $\hat{y}$.
\ENDFOR
\end{algorithmic}
\end{algorithm}

\section{Description of the Benchmark Datasets}\label{sec:benchmarks}

We evaluate on the test sets of six  arithmetic reasoning benchmarks. Two datasets include simpler problems that can be solved mostly in a single step: AddSub~\citep{hosseini2014} consists of 395 math word problems that require addition and / or subtraction for the solution, while SingleEQ~\citep{koncel-kedziorski2015} contains 508 questions which can be solved using a single equation. Four more challenging datasets require multi-step reasoning: MultiArith~\citep{roy2015} (600 math problems), SVAMP~\citep{patel2021} (1000 varied math problems), GSM8K~\citep{cobbe2021gsm8k} (1319 grade-school level problems), and AQuA~\citep{ling2017} (254 algebraic word problems). Although these arithmetic problems in the previous benchmarks are relatively simple for humans, LLMs often struggle in solving these types of problems~\citep{patel2021}. In addition, we also conduct experiments on considerably harder MATH~\citep{hendrycksmath2021} dataset which contains 5000 competition-level mathematics problems written in LaTeX and natural language. 
BIG-Bench~Hard~\citep{suzgun2023bbh} 
consists of 23 difficult tasks from the BIG-Bench suite \citep{srivastava2023beyond}, where previous large language models did not surpass the average human performance. We focus on the ``Date Understanding" and ``Object Tracking" tasks, which require quantitative reasoning.
Answering questions from the Date Understanding dataset involves  inferring a date from a given scenario.
Object tracking task evaluates an algorithm's ability to reason and determine the final state of objects, after applying a sequence of shuffling, starting from their known initial states. 
All of these benchmarks are available under open-source licenses~\footnote{CC-BY-4.0 [AddSub; SingleEQ], Apache 2.0 [MultiArith; AQuA] and MIT [SVAMP; GSM8K; MATH; BIG-Bench Hard].}.

\section{Experimental Results using Llama Models}
\label{sec:llama}
We have conducted experiments with two Llama-family LLMs: the weaker Llama-3-8b-instruct and the very capable Llama-3-70b-instruct.
In order to reduce the API cost of the experiments, we restrict running the more expensive 70B model to only the three most difficult benchmarks.

From the results in Table~\ref{tab:result_llama}, we observe that using Llama-3-8b-instruct, the relative advantage of PHP over CoT is diminished in comparison to the GPT models.
This suggests that weaker LLMs, such as Llama-3-8b-instruct, which often have relatively poor instruction following capability, cannot utilize the hint effectively for solving the reasoning task, highlighting the inadequacy of sophisticated prompting for weaker LLMs. In this setting, the effect of the quality of approximation of the initial distribution of RAD becomes important for obtaining a good reasoning accuracy and PHP+RAD outperforms CoT+RAD in most cases. Except for GSM8K, PHP+RAD either outperforms CoT+SC or obtains comparable performance on all other datasets. On the contrary, for a strongly capable Llama-3-70b-instruct model, both CoT+RAD and PHP+RAD perform well.
\begin{table}
\caption{Mean and standard error of accuracy (in \%) of few-shot arithmetic reasoning. The \textbf{highest} accuracy among all competing algorithms using the same LLM is marked in \textbf{bold} and is shown in {\color{red}\textbf{red}} and {\color{blue}\textbf{blue}} for {\color{red}\textbf{Llama-3-8b-instruct}} and  {\color{blue}\textbf{Llama-3-70b-instruct}} respectively. 
The \underline{second-best} accuracy in those cases is marked with an \underline{underline} and is shown in \underline{{\color{light_red}light red}} and \underline{{\color{light_blue}light blue}} respectively. 
}
\label{tab:result_llama}
\centering
\scriptsize
\begin{tabular}{cccccccc}
\toprule 
\textbf{LLM} &\textbf{Algorithm}   &\textbf{AddSub}        &\textbf{MultiArith}         &\textbf{SingleEQ} &\textbf{SVAMP}        &\textbf{GSM8K}         &\textbf{AQuA} \\ \midrule[0.25ex]
{\multirow{5}{*}{\rotatebox[origin=c]{90}{{\color{red}\textbf{\begin{tabular}[c]{@{}c@{}}Llama-3-8b-\\ instruct\end{tabular}}}}}} 

&\textbf{CoT}            &88.9$\pm$1.6 &96.7$\pm$0.7 &90.0$\pm$1.3 &83.5$\pm$1.2 &76.6$\pm$1.2 &51.2$\pm$3.1   \\

&\textbf{PHP}   &90.4$\pm$1.5 &94.7$\pm$0.9 &91.1$\pm$1.3 &86.4$\pm$1.1 &76.8$\pm$1.2 &57.1$\pm$3.1   \\

&\textbf{CoT+SC}         &\underline{\color{light_red}91.1$\pm$1.4} &{\color{red}\textbf{98.0$\pm$0.6}}$^*$ &94.5$\pm$1.0 &{\color{red}\textbf{90.4$\pm$0.9}}$^*$ &{\color{red}\textbf{85.0$\pm$1.0}}$^*$ &59.4$\pm$3.1   \\


\cmidrule[0.025ex]{2-8}
&\textbf{CoT+RAD}  &{\color{red}\textbf{92.9$\pm$1.3}}$^*$ &96.8$\pm$0.7 &\underline{\color{light_red}94.9$\pm$1.0} &\underline{\color{light_red}90.1$\pm$0.9} &82.3$\pm$1.1 &\underline{\color{light_red}60.0$\pm$3.0}    \\
&\textbf{PHP+RAD}  &{\color{red}\textbf{92.9$\pm$1.3}}$^*$ &\underline{\color{light_red}97.8$\pm$0.6} &{\color{red}\textbf{95.1$\pm$1.0}}$^*$ &{\color{red}\textbf{90.4$\pm$0.9}}$^*$ &\underline{\color{light_red}84.2$\pm$1.1} &{\color{red}\textbf{66.1$\pm$3.0}}$^*$  \\
\midrule[0.25ex] 
\midrule[0.25ex]
{\multirow{5}{*}{\rotatebox[origin=c]{90}{{\color{blue}\textbf{\begin{tabular}[c]{@{}c@{}}Llama-3-70b-\\ intruct\end{tabular}}}}}} 

&\textbf{CoT}            &- &- &- &91.2$\pm$0.9 &93.2$\pm$0.7 &72.8$\pm$2.8  \\

&\textbf{PHP}   &- &- &- &91.9$\pm$0.9 &93.3$\pm$0.7 &73.2$\pm$2.8  \\

&\textbf{CoT+SC}         &- &- &- &92.6$\pm$0.8 &\underline{\color{light_blue}94.2$\pm$0.6} &78.0$\pm$2.6   \\


\cmidrule[0.025ex]{2-8}
&\textbf{CoT+RAD}  &- &- &- &{\color{blue}\textbf{93.1$\pm$0.8}}$^*$ &\underline{\color{light_blue}94.2$\pm$0.6} &{\color{blue}\textbf{79.9$\pm$2.5}}$^*$   \\
&\textbf{PHP+RAD}  &- &- &- &\underline{\color{light_blue}92.7$\pm$0.8} &{\color{blue}\textbf{94.6$\pm$0.6}}$^*$ &\underline{\color{light_blue}78.7$\pm$2.6}  \\
\bottomrule
\end{tabular}
\end{table}

\section{Results for the `Difficult' Questions}
In order to demonstrate the advantage of CoT+RAD more clearly, we restrict
ourselves to only the `difficult’ questions in the six arithmetic benchmarks.
If a question is solved correctly by all
algorithms in Table~\ref{tab:result_difficult}, we categorize it as `easy’.
A question which is not `easy’ is termed `difficult’.
All easy questions are
subsequently removed from the datasets to compute the accuracies only
on the difficult questions.
From Table~\ref{tab:result_difficult}, we observe that the relative accuracy gains offered by the proposed CoT+RAD algorithm
are more substantial in most cases.
\begin{table}[htbp!]
\vspace{-1em}
\caption{Mean and standard error of accuracy (in \%) of few-shot arithmetic reasoning for the `difficult' questions. The \textbf{highest} accuracy among all competing algorithms using the same LLM is marked in \textbf{bold} and is shown in {\color{red}\textbf{red}}, {\color{blue}\textbf{blue}}, and {\color{orange}\textbf{orange}} for {\color{red}\textbf{GPT-3.5 Turbo}}, {\color{blue}\textbf{GPT-4 Turbo}}, and {\color{orange}\textbf{GPT-4o-mini}} respectively. 
The \underline{second-best} accuracy in those cases is marked with an \underline{underline} and is shown in \underline{{\color{light_red}light red}}, \underline{{\color{light_blue}light blue}}, and \underline{{\color{light_orange}light orange}} respectively. The \textbf{highest} accuracy is marked with an asterisk if the difference from the \underline{second-best} accuracy is statistically significant.
}
\label{tab:result_difficult}
\centering
\scriptsize
\begin{tabular}{cccccccc}
\toprule 
\textbf{LLM} &\textbf{Algorithm}   &\textbf{AddSub}        &\textbf{MultiArith}         &\textbf{SingleEQ} &\textbf{SVAMP}        &\textbf{GSM8K}         &\textbf{AQuA} \\ \midrule[0.25ex]
{\multirow{5}{*}{\rotatebox[origin=c]{90}{{\color{red}\textbf{GPT-3.5 Turbo}}}}}

&\textbf{CoT}  &\underline{{\color{light_red}46.0$\pm$6.3}} &51.9$\pm$9.6 &73.7$\pm$5.8 &36.5$\pm$2.9 &37.1$\pm$2.2 &30.7$\pm$3.7 \\

&\textbf{PHP}   &{\color{red}\textbf{47.6$\pm$6.2}}$^*$ &\underline{{\color{light_red}81.5$\pm$7.4}} &\underline{{\color{light_red}78.9$\pm$5.4}} &41.8$\pm$2.9 &51.3$\pm$2.3 &32.0$\pm$3.7 \\

&\textbf{CoT+SC}   &44.4$\pm$6.3 &77.8$\pm$7.9 &\underline{{\color{light_red}78.9$\pm$5.4}} &\underline{{\color{light_red}47.7$\pm$3.0}} &51.3$\pm$2.3 &\underline{{\color{light_red}49.0$\pm$4.0}} \\ 

&\textbf{PHP+SC}     &41.3$\pm$6.3 &74.1$\pm$8.4 &77.2$\pm$5.6 &41.4$\pm$2.9 &\underline{{\color{light_red}57.2$\pm$2.3}} &40.5$\pm$4.0 \\
\cmidrule[0.025ex]{2-8}
&\textbf{CoT+RAD}  &{\color{red}\textbf{47.6$\pm$6.3}}$^*$ &{\color{red}\textbf{92.6$\pm$5.1}}$^*$ &\textbf{{\color{red}82.5$\pm$5.1}}$^*$ &{\color{red}\textbf{51.6$\pm$3.0}}$^*$ &\textbf{{\color{red}63.8$\pm$2.2}}$^*$ &{\color{red}\textbf{51.0$\pm$4.0}}$^*$ \\
\midrule[0.25ex] \midrule[0.25ex]

{\multirow{5}{*}{\rotatebox[origin=c]{90}{{\color{blue}\textbf{GPT-4 Turbo}}}}}

&\textbf{CoT}  &{\color{blue}\textbf{77.8$\pm$5.3}} &63.0$\pm$9.3 &68.4$\pm$6.2 &73.0$\pm$2.6 &60.7$\pm$2.3 &73.2$\pm$3.6 \\

&\textbf{PHP}   &{\color{blue}\textbf{77.8$\pm$5.3}} &\underline{{\color{light_blue}66.7$\pm$9.2}} &\underline{{\color{light_blue}77.2$\pm$5.5}} &76.5$\pm$2.5 &\underline{{\color{light_blue}75.2$\pm$2.0}} &73.2$\pm$3.6 \\

&\textbf{CoT+SC}   &\underline{{\color{light_blue}76.2$\pm$5.4}} &{\color{blue}\textbf{74.1$\pm$8.4}}$^*$ &73.7$\pm$5.8 &76.8$\pm$2.5 &66.7$\pm$2.2 &{\color{blue}\textbf{76.5$\pm$3.5}}$^*$ \\ 

&\textbf{PHP+SC} 
&74.6$\pm$5.4 &{\color{blue}\textbf{74.1$\pm$8.5}}$^*$ &71.9$\pm$5.9 &\underline{{\color{light_blue}78.6$\pm$2.4}} &74.3$\pm$2.1 &71.2$\pm$3.6 \\
\cmidrule[0.025ex]{2-8}
&\textbf{CoT+RAD}  
&{\color{blue}\textbf{77.8$\pm$5.3}} &{\color{blue}\textbf{74.1$\pm$8.4}}$^*$ &{\color{blue}\textbf{87.7$\pm$4.4}}$^*$ &{\color{blue}\textbf{81.1$\pm$2.3}}$^*$ &{\color{blue}\textbf{84.4$\pm$1.7}}$^*$ &\underline{{\color{light_blue}73.9$\pm$3.5}} \\ 
\midrule[0.25ex] 
\midrule[0.25ex]

{\multirow{5}{*}{\rotatebox[origin=c]{90}{{\color{orange}\textbf{GPT-4o-mini}}}}} 

&\textbf{CoT}            &55.6$\pm$6.2 &74.1$\pm$8.5 &50.9$\pm$6.6 &77.2$\pm$2.5 &75.4$\pm$2.0 &64.7$\pm$3.9   \\

&\textbf{PHP}   &61.9$\pm$6.2 &74.1$\pm$8.4 &\underline{{\color{light_orange}57.9$\pm$6.6}} &77.5$\pm$2.5 &80.3$\pm$1.9 &64.7$\pm$3.8   \\

&\textbf{CoT+SC}         &55.6$\pm$6.3 &74.1$\pm$8.4 &56.1$\pm$6.6 &\underline{{\color{light_orange}78.9$\pm$2.4}} &\underline{{\color{light_orange}81.6$\pm$1.8}} &71.2$\pm$3.7  \\

&\textbf{PHP+SC}        &55.6$\pm$6.3 &74.1$\pm$8.5 &56.1$\pm$6.5 &76.8$\pm$2.5 &80.9$\pm$1.9 &\underline{{\color{light_orange}73.9$\pm$3.5}}   \\

\cmidrule[0.025ex]{2-8}
&\textbf{CoT+RAD}  &\textbf{{\color{orange}65.1$\pm$6.1}}$^*$ &74.1$\pm$8.4 &\textbf{{\color{orange}61.4$\pm$6.4}}$^*$ &\textbf{{\color{orange}79.3$\pm$2.4}}$^*$ &{\color{orange}\textbf{83.6$\pm$1.7}}$^*$ &\textbf{{\color{orange}74.5$\pm$3.5}}$^*$   \\
\bottomrule
\end{tabular}
\end{table}



\section{Additional Results for Comparing Probability of Correct Answer}
\label{sec:prob_other_llm}
\begin{figure}[ht]
\centering
\includegraphics[trim={0em 33em 0em 4em }, clip, width=\textwidth]{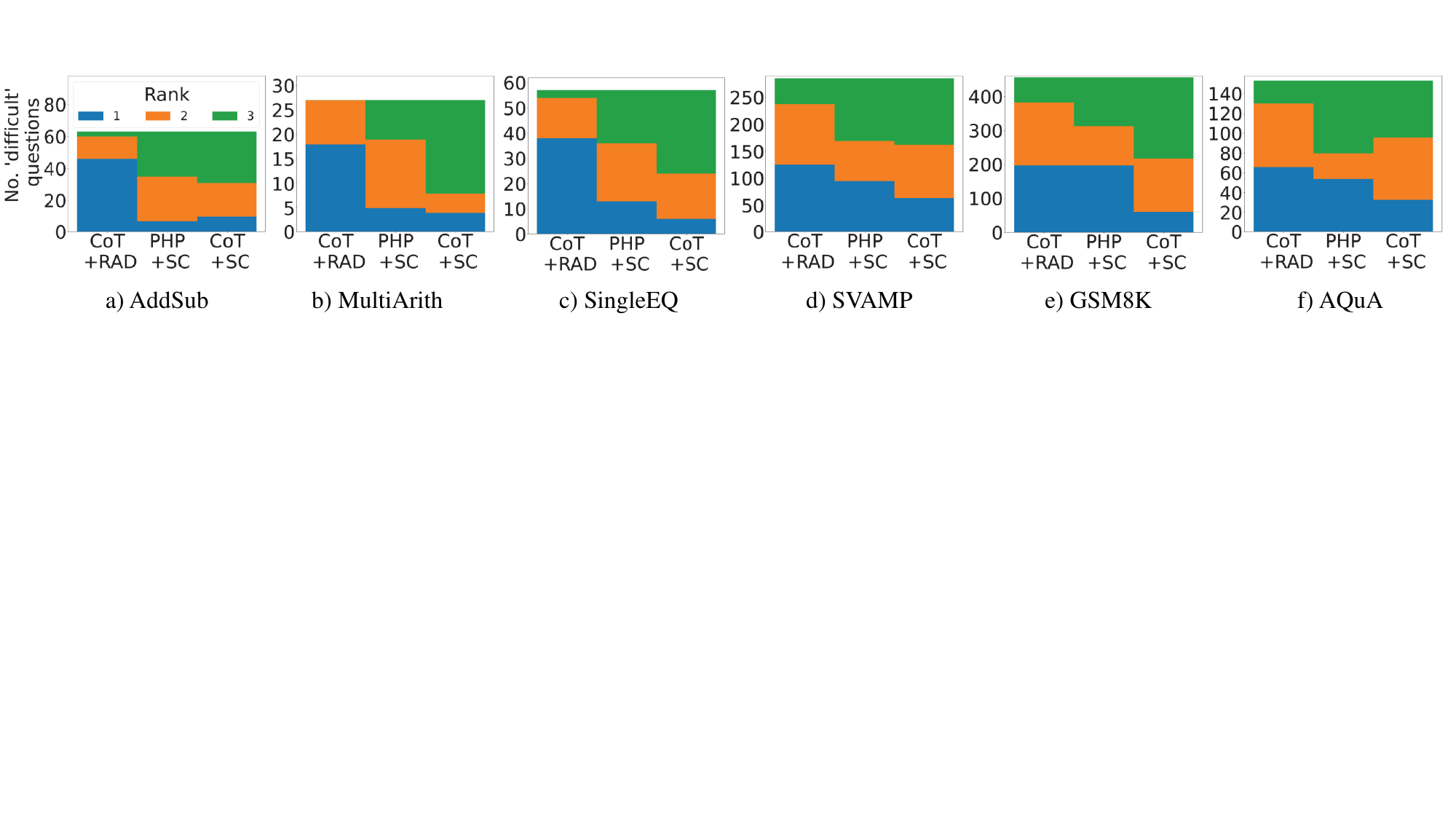}
\caption{Histogram of \textbf{ranks} of the algorithms (the \textbf{highest probability of the correct answer} results in the \textbf{lowest rank}) for the \textbf{`difficult'} questions from all six arithmetic datasets using \textbf{GPT-3.5 Turbo}.}
\label{fig:gpt3.5-stacked}
\end{figure}

\begin{figure}
\centering
\includegraphics[trim={0em 33em 0em 4em }, clip, width=\textwidth]{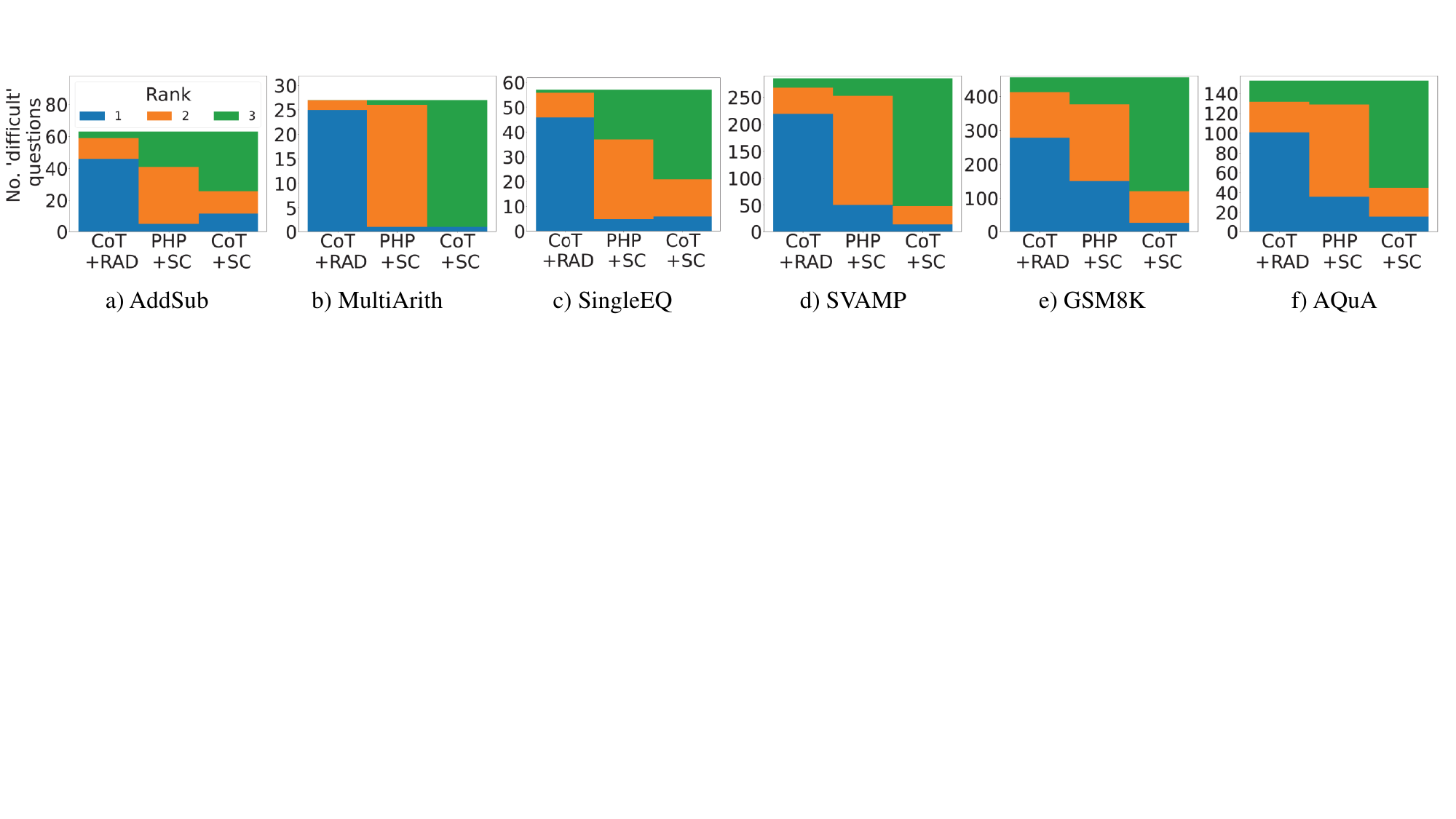}
\caption{Histogram of \textbf{ranks} of the algorithms (the \textbf{highest probability of the correct answer} results in the \textbf{lowest rank}) for the \textbf{`difficult'} questions from all six arithmetic datasets using \textbf{GPT-4 Turbo}.}
\label{fig:gpt4-stacked}
\end{figure}

Figure~\ref{fig:gpt4o-mini-stacked} in the main paper shows that in comparison to CoT+SC and PHP+SC using GPT-4o-mini, CoT+RAD assigns higher probability to the correct answers for most of the `difficult' questions across all datasets. 
Figures~\ref{fig:gpt3.5-stacked} and~\ref{fig:gpt4-stacked} demonstrate that the same trend holds for both GPT-3.5-Turbo and GPT-4-Turbo LLMs.

In order to demonstrate the statistical significance of the increase in probability of the true answer, we conduct a Wilcoxon signed rank test between $p_3(y|x)$ (i.e., the estimated probability of the true answer obtained from the proposed CoT+RAD) and $p_1(y|x)$ (i.e., the probability of the true answer, at the initialization of CoT+RAD, estimated from CoT+SC using 40 samples), and report the p-values in Table~\ref{tab:p_values}.
We observe that except for 5 out of 36 cases (6 datasets, 3 LLMs, and 2 different partitions of the datasets), the difference between $p_3(y|x)$ and $p_1(y|x)$ is statistically significant at the 5\% level, providing strong empirical support in favor of the capability of the RAD iterations in increasing the probability of the true answers.

\begin{table}[h]
\vspace{-0.5em}
\caption{p-value from Wilcoxon signed rank test between the probabilities of true answers from distributions $p_3(y|x)$ and $p_1(y|x)$ for the `difficult' questions (for the entire dataset)}
\label{tab:p_values}
\centering
\resizebox{0.975\columnwidth}{!}{
\scriptsize
\begin{tabular}{cccccccc}
\toprule 
\textbf{LLM}           & \textbf{AddSub} & \textbf{MultiArith} & \textbf{SingleEQ} & \textbf{SVAMP} & \textbf{GSM8K} & \textbf{AQuA} \\ \bottomrule
\textbf{GPT-3.5 Turbo} &0.0291 (0.1172)     &0.0006 ($1.3{\times} 10^{-5}$)           &0.0012 ($8.6{\times}10^{-5}$)         &0.0132 ($1.4{\times} 10^{-5}$)       &$9.2 {\times} 10^{-18}$ ($4.3{\times}10^{-22}$)       &0.0001  ($1.6{\times} 10^{-8}$)    \\ 
\textbf{GPT-4 Turbo}              &0.2868 (0.2258)        &0.0104 ($2.3{\times} 10^{-6}$)            &0.0002 ($6.2 {\times} 10^{-7}$)          &$4.8 {\times} 10^{-8}$ ($1.7{\times} 10^{-13}$)      &$2.2 {\times} 10^{-31}$ ($1.5{\times} 10^{-41}$)      &0.0065 (0.0042)     \\ 
\textbf{GPT-4o-mini}              &0.0038 (0.0024)      &0.8413 (0.0243)           &0.0317 (0.0255)          &0.5898 (0.3028)       &$4.5{\times} 10^{-12}$ ($5.2{\times} 10^{-12}$)       &$2.1{\times} 10^{-5}$ ($8.5 {\times} 10^{-6}$)     \\ \bottomrule
\end{tabular}}
\end{table}

In addition, we also calculate the percentage of difficult questions for which $p_3(y|x) \geqslant p_1(y|x)$ is satisfied and report the results in Table~\ref{tab:percent_of_inc_prob}.
We observe that in each case, for the majority of the questions, RAD iterations do not decrease the probability of the true answer.

\begin{table}[h!]
\vspace{-0.5em}
\caption{Percentage of `difficult' questions (percentage of questions in the entire dataset), so that $p_3(y|x) \geqslant p_1(y|x)$ is satisfied (in other words, RAD does not decrease the probability of the true answer)}
\label{tab:percent_of_inc_prob}
\centering
\scriptsize
\begin{tabular}{cccccccc}
\toprule 
\textbf{LLM}           & \textbf{AddSub} & \textbf{MultiArith} & \textbf{SingleEQ} & \textbf{SVAMP} & \textbf{GSM8K} & \textbf{AQuA} \\ \bottomrule
\textbf{GPT-3.5 Turbo} &79.4 (92.7)        &85.2 (97.3)           &86.0 (97.2)        &63.5 (83.8)       &70.8 (81.4)      &64.7 (74.8)      \\ 
\textbf{GPT-4 Turbo}              &76.2 (95.7)        &96.3 (99.7)          &87.7 (98.0)          &89.5 (96.9)     &85.7 (93.3)       &79.1 (86.6)     \\ 
\textbf{GPT-4o-mini}              &85.7 (97.2)       &96.3 (99.7)           &82.5  (97.0)          &81.1 (93.9)      &83.8 (92.7)       &75.8 (83.9)    \\ \bottomrule
\end{tabular}
\end{table}

\section{Refinement Prompt using Hinting~\citep{zheng2023progressive}}
\label{sec:hint-prompt}

A typical arithmetic reasoning question is presented in Table \ref{tab:example}, where the chain-of-thought yielded the incorrect answer $y=21$. In Table \ref{tab:hinting-prompt}, the hinting-based refinement prompt of PHP is applied to this same question with the hints $y'=4, 7$, yielding a chain-of-thought with the correct answer $y=6$.

\begin{table}[htbp!]
\centering
\caption{Example problem from \citet{zheng2023progressive}.}
\label{tab:example}
\vspace{5pt}
\begin{tabularx}{\linewidth}{>{\raggedright\arraybackslash}X}
\toprule
Q: There are 15 trees in the grove. Grove workers will plant trees in the grove today. After they are done, there will be 21 trees. How many trees did the grove workers plant today?

\\
A: There are 21 trees in the Grove today. Grove workers are done planting trees. So 21 trees were planted in the Grove. 
\\\bottomrule
\end{tabularx}
\end{table}

\begin{table}[htbp!]
\centering
\caption{Demonstration of the hinting prompt from \citet{zheng2023progressive}, as applied to the example problem from Table \ref{tab:example}. Additions are highlighted in blue.}
\label{tab:hinting-prompt}
\vspace{5pt}
\begin{tabularx}{\linewidth}{>{\raggedright\arraybackslash}X}
\toprule
Q: There are 15 trees in the grove. Grove workers will plant trees in the grove today. After they are done, there will be 21 trees. How many trees did the grove workers plant today? \textcolor{blue}{(Hint: The answer is near to \textcolor{red}{4, 7}).}

\\
A: \textcolor{blue}{We know the Answer Hints: \textcolor{red}{4, 7}. With the Answer Hints: \textcolor{red}{4, 7}, we will answer the question.} There are
15 trees originally. Then there were 21 trees after the Grove workers planted some more. So there
must have been 21 - 15 = 6 trees that were planted. The answer is 6. 
\\\bottomrule
\end{tabularx}
\end{table}

\section{Examples of Full Prompts}\label{sec:full-prompts}

We provide examples of full prompts for CoT in Table~\ref{tab:cot-example}, PHP in Table~\ref{tab:php-example} and our method CoT+RAD in Table~\ref{tab:cot+hm-example}. As CoT+SC and PHP+SC are simply sampled versions of CoT and PHP, the prompts are identical as in those cases. The base examples are taken from \citet{zheng2023progressive}.

\begin{table}[h!]
\centering
\caption{An example of a CoT prompt with 4 few-shot examples.}
\label{tab:cot-example}
\begin{tabularx}{\linewidth}{>{\raggedright\arraybackslash}X}
\toprule
Q: There are 15 trees in the grove. Grove workers will plant trees in the grove today. After they are
done, there will be 21 trees. How many trees did the grove workers plant today?

A: There are 15 trees originally.Then there were 21 trees after the Grove workers planted some more.
So there must have been 21 - 15 = 6 trees that were planted. The answer is 6.

\\
Q: If there are 3 cars in the parking lot and 2 more cars arrive, how many cars are in the parking lot?

A: There are originally 3 cars. Then 2 more cars arrive. Now 3 + 2 = 5 cars are in the parking lot.
The answer is 5.

\\
Q: Leah had 32 chocolates and her sister had 42. If they ate 35, how many pieces do they have left in
total?

A: Originally, Leah had 32 chocolates and her sister had 42. So in total they had 32 + 42 = 74. After
eating 35, they had 74 - 35 = 39 pieces left in total. The answer is 39.

\\
Q: Jason had 20 lollipops. He gave Denny some lollipops. Now Jason has 12 lollipops. How many
lollipops did Jason give to Denny?

A: Jason had 20 lollipops originally. Then he had 12 after giving some to Denny. So he gave Denny
20 - 12 = 8 lollipops. The answer is 8.

\\
Q: Shawn has five toys. For Christmas, he got two toys each from his mom and dad. How many toys
does he have now?

A: 
%
\\\bottomrule
\end{tabularx}
\end{table}

\begin{table}[ht!]
\centering
\caption{An example of a PHP prompt with 4 few-shot examples. The few-shot questions are provided true answer, while the test question is provided with the history so far, in this case $y'=7, 11, 8$.}
\label{tab:php-example}
\vspace{5pt}
\begin{tabularx}{\linewidth}{>{\raggedright\arraybackslash}X}
\toprule
Q: There are 15 trees in the grove. Grove workers will plant trees in the grove today. After they are
done, there will be 21 trees. How many trees did the grove workers plant today? \textcolor{blue}{(Hint: The answer is near to \textcolor{red}{6}).}

A: \textcolor{blue}{We know the Answer Hints: \textcolor{red}{6}. With the Answer Hints: \textcolor{red}{6}, we will answer the question.} There are 15 trees originally. Then there were 21 trees after the Grove workers planted some more.
So there must have been 21 - 15 = 6 trees that were planted. The answer is 6.

\\
Q: If there are 3 cars in the parking lot and 2 more cars arrive, how many cars are in the parking lot? \textcolor{blue}{(Hint: The answer is near to \textcolor{red}{5}).}

A: \textcolor{blue}{We know the Answer Hints: \textcolor{red}{5}. With the Answer Hints: \textcolor{red}{5}, we will answer the question.} There are originally 3 cars. Then 2 more cars arrive. Now 3 + 2 = 5 cars are in the parking lot.
The answer is 5.

\\
Q: Leah had 32 chocolates and her sister had 42. If they ate 35, how many pieces do they have left in
total? \textcolor{blue}{(Hint: The answer is near to \textcolor{red}{39}).}

A: \textcolor{blue}{We know the Answer Hints: \textcolor{red}{39}. With the Answer Hints: \textcolor{red}{39}, we will answer the question.} Originally, Leah had 32 chocolates and her sister had 42. So in total they had 32 + 42 = 74. After
eating 35, they had 74 - 35 = 39 pieces left in total. The answer is 39.

\\
Q: Jason had 20 lollipops. He gave Denny some lollipops. Now Jason has 12 lollipops. How many
lollipops did Jason give to Denny? \textcolor{blue}{(Hint: The answer is near to \textcolor{red}{8}).}

A: \textcolor{blue}{We know the Answer Hints: \textcolor{red}{8}. With the Answer Hints: \textcolor{red}{8}, we will answer the question.} Jason had 20 lollipops originally. Then he had 12 after giving some to Denny. So he gave Denny
20 - 12 = 8 lollipops. The answer is 8.

\\
Q: Shawn has five toys. For Christmas, he got two toys each from his mom and dad. How many toys
does he have now? \textcolor{blue}{(Hint: The answer is near to \textcolor{red}{7, 11, 8}).}

A: 
%
\\\bottomrule
\end{tabularx}
\end{table}

\begin{table}[ht]
\centering
\caption{An example of a CoT+RAD prompt with 4 few-shot examples. The few-shot questions are provided with true answers as hints, while the test question is provided with one of the distinct answers obtained during the interaction with the LLM in the previous iteration. In this case, $y'=8$.}
\label{tab:cot+hm-example}
\vspace{5pt}
\begin{tabularx}{\linewidth}{>{\raggedright\arraybackslash}X}
\toprule
Q: There are 15 trees in the grove. Grove workers will plant trees in the grove today. After they are
done, there will be 21 trees. How many trees did the grove workers plant today? \textcolor{blue}{(Hint: The answer is near to \textcolor{red}{6}).}

A: \textcolor{blue}{We know the Answer Hints: \textcolor{red}{6}. With the Answer Hints: \textcolor{red}{6}, we will answer the question.} There are 15 trees originally. Then there were 21 trees after the Grove workers planted some more.
So there must have been 21 - 15 = 6 trees that were planted. The answer is 6.

\\
Q: If there are 3 cars in the parking lot and 2 more cars arrive, how many cars are in the parking lot? \textcolor{blue}{(Hint: The answer is near to \textcolor{red}{5}).}

A: \textcolor{blue}{We know the Answer Hints: \textcolor{red}{5}. With the Answer Hints: \textcolor{red}{5}, we will answer the question.} There are originally 3 cars. Then 2 more cars arrive. Now 3 + 2 = 5 cars are in the parking lot.
The answer is 5.

\\
Q: Leah had 32 chocolates and her sister had 42. If they ate 35, how many pieces do they have left in
total? \textcolor{blue}{(Hint: The answer is near to \textcolor{red}{39}).}

A: \textcolor{blue}{We know the Answer Hints: \textcolor{red}{39}. With the Answer Hints: \textcolor{red}{39}, we will answer the question.} Originally, Leah had 32 chocolates and her sister had 42. So in total they had 32 + 42 = 74. After
eating 35, they had 74 - 35 = 39 pieces left in total. The answer is 39.

\\
Q: Jason had 20 lollipops. He gave Denny some lollipops. Now Jason has 12 lollipops. How many
lollipops did Jason give to Denny? \textcolor{blue}{(Hint: The answer is near to \textcolor{red}{8}).}

A: \textcolor{blue}{We know the Answer Hints: \textcolor{red}{8}. With the Answer Hints: \textcolor{red}{8}, we will answer the question.} Jason had 20 lollipops originally. Then he had 12 after giving some to Denny. So he gave Denny
20 - 12 = 8 lollipops. The answer is 8.

\\
Q: Shawn has five toys. For Christmas, he got two toys each from his mom and dad. How many toys
does he have now? \textcolor{blue}{(Hint: The answer is near to \textcolor{red}{8}).}

A: 
%
\\\bottomrule
\end{tabularx}
\end{table}





\end{document}